%% file: main.tex
\documentclass[10pt]{article}

\usepackage[]{berkeley}
\input{preamble}

\usepackage{booktabs,tabularx}
\usepackage{array}          %
\usepackage{pifont}         %
\newcommand{\cmark}{\ding{51}}

\newcolumntype{Y}{>{\centering\arraybackslash}X}

\usepackage{xspace}
\newcommand{\ours}{\textsc{VisGym}\xspace}
\def\paperID{266} %
\def\confName{CVPR}
\def\confYear{2026}

\title{\ours:
Diverse, Customizable, Scalable\\Environments for Multimodal Agents}

\author{
{\normalsize
\textbf{Zirui Wang}$^{\dagger}$,
\textbf{Junyi Zhang}$^{\dagger}$,
\textbf{Jiaxin Ge}$^{\dagger}$,
\textbf{Long Lian},
\textbf{Letian Fu},
\textbf{Lisa Dunlap},\\
\textbf{Ken Goldberg},
\textbf{XuDong Wang},
\textbf{Ion Stoica},
\textbf{David M. Chan},
\textbf{Sewon Min},
\textbf{Joseph E. Gonzalez}
}\\[0.5em]
\normalsize\textbf{UC Berkeley} \hfill {\small $^{\dagger}$Equal contribution.}
}

\begin{document}

\maketitle

\input{sec/1_intro}

\input{sec/2_gym}

\input{sec/3_evaluation}

\input{sec/4_diagnosis}

\input{sec/5_training}

\input{sec/6_related_work}

\input{sec/7_conclusion}

{
    \setcitestyle{numbers,sort&compress}
    \bibliographystyle{plainnat}
    \bibliography{main}
}

\input{sec/8_appendix}

\end{document}

%% file: preamble.tex
\usepackage{silence}
\usepackage{etoc}
\usepackage{newtxtext}
\usepackage{xspace}
\usepackage{makecell}
\usepackage{adjustbox}
\usepackage{amsmath}
\usepackage{enumitem}
\setlist[itemize]{label=\textbullet, leftmargin=*}
\usepackage{stfloats}

\usepackage[noend]{algpseudocode}
\usepackage{microtype}
\usepackage{graphicx}
\usepackage{xcolor}
\usepackage{paralist}
\usepackage{mathtools}
\usepackage{algorithm}
\usepackage{tikz}
\usetikzlibrary{arrows,automata,positioning}
\usepackage{multirow} 
\usepackage{xspace}
\usepackage{comment}
\usepackage{amssymb} %
\makeatletter
\renewcommand*\PackageWarning[2]{%
  \ifx#1multicol\relax\else
    \GenericWarning{}{#1: #2}%
  \fi}
\makeatother
\newcommand{\EnvEntry}[2]{%
  \noindent\hyperref[#2]{#1}\dotfill\pageref{#2}\par
  \vspace{-1ex}
}
\usepackage{dsfont}
\usepackage{tabulary}
\usepackage{tabularx}
\usepackage{cuted}
\usepackage{wrapfig}

\usepackage{dsfont}
\usepackage{tabulary}
\usepackage{wrapfig}
\usepackage{xspace}
\usepackage{mathtools}
\usepackage{amsthm}
\usepackage{array}
\usepackage{tcolorbox}
\tcbuselibrary{skins,breakable}
\usepackage{needspace}
\definecolor{berkeleyblue}{HTML}{002676}

\usepackage[
    pagebackref,
    breaklinks,
    colorlinks,
    linkcolor=berkeleyblue,
    citecolor=berkeleyblue,
    urlcolor=berkeleyblue,
    filecolor=berkeleyblue
]{hyperref}

\newcolumntype{L}[1]{>{\raggedright\arraybackslash}m{#1}}
\newcolumntype{S}[1]{>{\raggedright\arraybackslash}p{#1}}

\usepackage{lineno}

\newif\ifshowcomments
\showcommentsfalse   %

\newenvironment{EnvCard}{%
\begin{tcolorbox}[
  enhanced,
  breakable,
  width=\linewidth,
  colback=envbg,
  colframe=envframe,
  coltitle=envtext,
  boxrule=0.6pt,
  left=1em,
  right=1em,
  top=0.5em,
  bottom=0.1em,
  rounded corners,
  borderline west={2pt}{0pt}{envaccent},
]
}%
{\end{tcolorbox}}

\ifshowcomments
  \newcommand{\colin}[1]{\textcolor{red}{[Colin: #1]}}
  \newcommand{\maxfu}[1]{\textcolor{blue}{[Max: #1]}}
  \newcommand{\jz}[1]{\textcolor{green}{[Junyi: #1]}}
  \newcommand{\david}[1]{\textcolor{magenta}{[David: #1]}}
  \newcommand{\jiaxin}[1]{\textcolor{pink}{[Jiaxin: #1]}}
\else
  \newcommand{\colin}[1]{}
  \newcommand{\maxfu}[1]{}
  \newcommand{\jz}[1]{}
  \newcommand{\david}[1]{}
  \newcommand{\jiaxin}[1]{}
\fi

\usepackage[margin=1in]{geometry} %
\usepackage{graphicx}      %
\usepackage{xcolor}        %
\usepackage{listings}      %
\usepackage{listingsutf8}

\usepackage{pgffor}        %
\usepackage{tikz}          %
\usepackage{adjustbox}     %
\newlength{\EnvStepCaseNeedspace}
\newcommand{\myparagraph}[1]{\smallskip\noindent\textbf{#1} }

\definecolor{lightyellow}{RGB}{225, 240, 250}
\definecolor{winered}{RGB}{160, 25, 45}

\definecolor{envbg}{RGB}{252,252,254}     %
\definecolor{envframe}{RGB}{210,210,220}  %
\definecolor{envaccent}{RGB}{0,47,167}       %
\definecolor{envtext}{RGB}{40,40,45}      %

\lstdefinestyle{envtxt}{
  basicstyle=\ttfamily\scriptsize\color{envtext},
  breaklines=true,
  breakautoindent=false,
  breakindent=0pt,
  frame=single,
  framerule=0.6pt,
  rulecolor=\color{envframe},
  backgroundcolor=\color{envbg},
  xleftmargin=0.6em,
  framexleftmargin=0.6em,
  frameshape={RYRY}{Y}{Y}{RYRY}, %
  moredelim=**[is][\color{envaccent}]{@}{@}, %
  columns=fullflexible,
  keepspaces=true,
}

\newcommand{\PromptLabel}[1]{%
  \ifnum#1=0
    \textbf{Instruction $I$}%
  \else
    \textbf{Feedback $f_{\number#1}$}%
  \fi
}

\newcommand{\ActionLabel}[1]{%
  \textbf{Action $a_{\number\numexpr#1+1\relax}$:}%
}

\newcommand{\EnvImage}[2]{%
  \noindent
  \centering
  \includegraphics[
    width=\linewidth,
    height=2.5cm,
    keepaspectratio
  ]{#1/image_#2.png}\par
}

\newcommand{\EnvTextFile}[3]{%
  \noindent
  \lstinputlisting[style=envtxt,inputencoding=utf8]{#1/#2_#3.txt}\par
}

\newcommand{\EnvStep}[2]{%
  \begin{EnvCard}

  \ifnum#2=0
      {\footnotesize \PromptLabel{#2}}\par
      \vspace{0.1em}
      \EnvTextFile{#1}{prompt}{#2}
      \vspace{0.1em}

      \noindent
      \begin{minipage}[t]{0.48\textwidth}
          {\footnotesize \textbf{Observation $o_{\number#2}$}}\par
          \vspace{0.5em}
          \EnvImage{#1}{#2}
      \end{minipage}
      \hfill %
      \begin{minipage}[t]{0.48\textwidth}
          {\footnotesize \ActionLabel{#2}}\par
          \vspace{0.1em}
          \EnvTextFile{#1}{action}{#2}
      \end{minipage}

  \else
      \noindent
      \begin{minipage}[t]{0.48\textwidth}
          {\footnotesize \textbf{Observation $o_{\number#2}$}}\par
          \vspace{0.5em}
          \EnvImage{#1}{#2}
      \end{minipage}
      \hfill
      \begin{minipage}[t]{0.48\textwidth}
          {\footnotesize \PromptLabel{#2}}\par
          \vspace{0.1em}
          \EnvTextFile{#1}{prompt}{#2}
          \vspace{0.1em}

          {\footnotesize \ActionLabel{#2}}\par
          \vspace{0.1em}
          \EnvTextFile{#1}{action}{#2}
      \end{minipage}
  \fi

  \end{EnvCard}
}

\newcommand{\EnvStepCase}[2]{%
  \Needspace{\EnvStepCaseNeedspace}
  \begin{EnvCard}

  \ifnum#2=0
      {\footnotesize \textbf{\textcolor{winered}{Reason}}}\par
      \vspace{0.1em}
      \noindent
      \lstinputlisting[
        style=envtxt,
        basicstyle=\ttfamily\bfseries\scriptsize\color{envtext}, 
        inputencoding=utf8
      ]{#1/reason.txt}\par
      \vspace{0.3em} %

      {\footnotesize \textbf{\textcolor{winered}{Evidence}}}\par
      \vspace{0.1em}
      \noindent
      \lstinputlisting[
        style=envtxt,
        basicstyle=\ttfamily\bfseries\scriptsize\color{envtext}, 
        inputencoding=utf8
      ]{#1/evidence.txt}\par
      \vspace{0.3em}

      {\footnotesize \PromptLabel{#2}}\par
      \vspace{0.1em}
      \EnvTextFile{#1}{prompt}{#2}
      \vspace{0.1em}

      \begin{minipage}[t]{0.48\textwidth}
          {\footnotesize \textbf{Observation $o_{\number#2}$}}\par
          \vspace{0.5em}
          \EnvImage{#1}{#2}
      \end{minipage}

      \vspace{0.1em}
      {\footnotesize \ActionLabel{#2}}\par
      \vspace{0.1em}
      \EnvTextFile{#1}{action}{#2}

      \vspace{0.1em}
      {\footnotesize \textbf{VLM Raw Output:}}\par
      \vspace{0.1em}
      \EnvTextFile{#1}{vlm_output}{#2}

  \else
      \noindent
      \begin{minipage}[t]{0.48\textwidth}
          {\footnotesize \textbf{Observation $o_{\number#2}$}}\par
          \vspace{0.5em}
          \EnvImage{#1}{#2}
      \end{minipage}
      \hfill
      \begin{minipage}[t]{0.48\textwidth}
          {\footnotesize \PromptLabel{#2}}\par
          \vspace{0.1em}
          \EnvTextFile{#1}{prompt}{#2}
          \vspace{0.1em}

          {\footnotesize \ActionLabel{#2}}\par
          \vspace{0.1em}
          \EnvTextFile{#1}{action}{#2}
      \end{minipage}
      
      \par 
      \vspace{0.5em} 
      
      {\footnotesize \textbf{VLM Raw Output:}}\par
      \vspace{0.1em}
      \EnvTextFile{#1}{vlm_output}{#2}
      
  \fi

  \end{EnvCard}
}

\newcommand{\EnvRollout}[4][]{%
  \clearpage
  \onecolumn
  \raggedbottom %
  \begingroup
    \if\relax\detokenize{#1}\relax
      \definecolor{envaccent}{RGB}{0,47,167}%
    \else
      \colorlet{envaccent}{#1}%
    \fi

    \refstepcounter{subsection}%
    \addcontentsline{toc}{subsection}{#2}%
    \vspace{0.5em}
    \noindent
    {\large \thesubsection\quad \textbf{#2}\quad
      \normalsize(\hyperref[appx:task_details]{back to Table of Contents})%
    }%
    \label{appx:\thesubsection}%
    \vspace{0.5em}

    \noindent
    \EnvStep{#3}{0}%
    \par\vspace{0.5em} %

    \foreach \i in {1,...,\numexpr#4-2\relax}{%
      \filbreak
      \noindent
      \EnvStep{#3}{\i}
      \par\vspace{0.5em} %
    }%
  \endgroup
  \clearpage
  \twocolumn
}

\newcommand{\EnvRolloutCase}[4][]{%
  \clearpage
  \onecolumn
  \raggedbottom %
  \begingroup
    \if\relax\detokenize{#1}\relax
      \definecolor{envaccent}{RGB}{0,47,167}%
    \else
      \colorlet{envaccent}{#1}%
    \fi

    \setcounter{subsection}{2}   %
    \renewcommand{\thesubsubsection}{F.2.\arabic{subsubsection}}
    \refstepcounter{subsubsection}%
    \addcontentsline{toc}{subsubsection}{#2}%
    \vspace{0.3em}
    \noindent
    {\large \thesubsubsection\quad \textbf{#2}\quad
      \normalsize(\hyperref[sec:stringsight_traj]{back to Sec. F.2})%
    }%
    \label{appx:\thesubsubsection}%
    \vspace{0.5em}

    \noindent
    \EnvStepCase{#3}{0}%
    \par\vspace{0.3em} %

    \foreach \i in {1,...,\numexpr#4-2\relax}{%
      \filbreak
      \noindent
      \EnvStepCase{#3}{\i}
      \par\vspace{0.3em} %
    }%
  \endgroup
  \clearpage
  \twocolumn
}

%% file: sec/1_intro.tex
\vspace{-4ex}
\begin{figure}[b]
    \centering
    \includegraphics[width=0.995\linewidth]{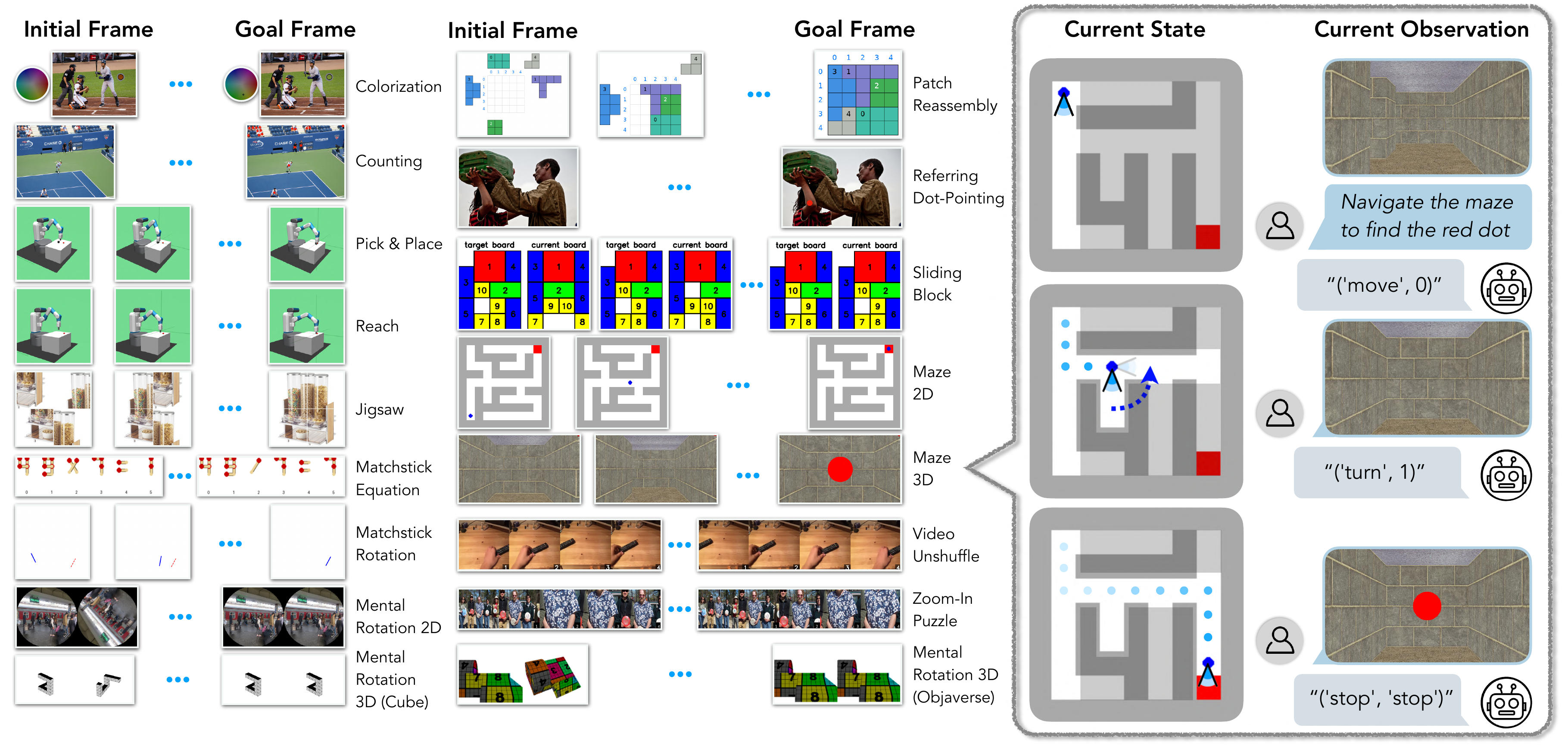}
    \vspace{-2ex}
    \captionsetup[figure]{hypcap=false}
    \caption{\textbf{An overview of \ours.} (\textit{Left}) \ours{} consists of $17$ diverse, long-horizon environments designed to systematically evaluate, diagnose, and train VLMs on visually interactive tasks with different domains, levels of state observability, and types of observations. (\textit{Right}) An example trajectory for the Maze 3D navigation task illustrates a partially observable environment consisting of non-structured synthetic renderings. Here, a VLM is prompted with (1) the task description (\textit{simplified in the figure}) and (2) a set of available actions to use (\textit{not shown in the figure for simplicity}). The agent must select each action conditioned on both its past actions and observation history for its decision-making. \colin{change h and t}}
    \label{fig:teaser}
\end{figure}

\input{tables/framework_comparisons}

\section{Introduction}
\label{sec:intro}

Humans navigate complex tasks in visually rich and interactive settings: manipulating objects, using devices, or exploring unfamiliar environments.
Success in these settings hinges on the tight coupling of perception, memory, and action over multiple steps \citep{Gibson1979TheEA, Henderson2001ASA}.
Foundation Vision–Language Models (VLMs) have made remarkable progress on static vision–language benchmarks
\citep{yue2024mmmu, lu2023mathvista, wang2024charxiv}
and on text-based multi-step tasks such as web browsing and coding
\citep{sirdeshmukh2025multichallenge, wei2025browsecomp, jimenez2023swe}.
Yet when visual observations must be integrated into multi-step decision-making, their behavior remains far less understood.
Recent evaluations across robotic manipulation, computer-use agents, and gaming agents highlight a range of challenges for visually interactive decision-making, including low task success rates, brittle visual grounding, and weak generalization \citep{zhang2025vlabench, xie2024osworld, zhang2025videogamebench, liu2023libero, hu2025lmgame, chen2025g1, shi2025korgym, liu2024visualagentbench}.
Although these insights are valuable, they tend to be domain-specific and observational, offering limited \emph{systematic, controlled} diagnosis of how domain-agnostic factors—such as context length, representation modality, feedback design, or goal visibility—affect model performance across tasks.

We introduce \ours{}, a highly diverse, scalable, and customizable gymnasium with 17 long-horizon environments designed to isolate what limits interactive decision-making across domains and to expose where current VLMs break down.
The suite spans symbolic puzzles, real-image understanding, navigation, and manipulation tasks, each with distinct observability and dynamics and equipped with oracle multi-step solvers for supervised finetuning (framework comparison in \cref{tab:framework_comparisons}).
Crucially, \ours{} provides fine-grained controls over input representation, difficulty, history length, planning horizon, and feedback, enabling domain-agnostic, systematic analysis of model behavior.
Building on prior domain-specific studies, we conduct cross-domain controlled experiments that examine how these factors, together with module finetuning and data curation, affect performance in multi-step visual decision-making.

Across $12$ state-of-the-art models, even the strongest achieve only $46.61\%$ and $26.00\%$ success in the easy and hard settings, respectively.
Our analyses reveal several concrete, cross-domain failure modes: 
\textbf{\textit{(1)}} models struggle to effectively leverage long-term context, showing a reversed-U relationship where performance degrades as the context grows unbounded; \textbf{\textit{(2)}} VLMs struggle with low-level perceptual grounding, a limitation highlighted by symbolic variants of tasks being substantially easier than their visually rendered counterparts; \textbf{\textit{(3)}} models struggle to infer task states and outcomes from purely visual transitions, consistently relying on explicit textual feedback to boost performance; \textbf{\textit{(4)}} the benefit of providing explicit goal observations is brittle and can backfire: while explicit goals can yield large gains, limited visual perception can cause models to misidentify them and, paradoxically, perform worse than with no goal at all; \textbf{\textit{(5)}} models fail to learn from standard demonstrations under partial observability or unknown dynamics, requiring information-revealing demonstrations that expose hidden states or clarify dynamics to significantly improve supervised finetuning outcomes.

Together, these findings establish \ours{} as a unified and extensible framework for diagnosing, understanding, and ultimately improving VLMs in visually interactive decision-making.

%% file: tables/framework_comparisons.tex
\begin{table*}[t]
\centering
\caption{\textbf{Comparison among frameworks for visually interactive decision-making.}
\textbf{Struct. Obs.} and \textbf{Non-struct. Obs.} indicate whether visual inputs can be parsed into structured text.
\textbf{POMDP} denotes partial observability with hidden states.
\textbf{Multi-Domain} covers diversity across domains (\textit{e.g.}, robotics, computer use, games, puzzles).
\textbf{Scalable Episodes} marks automatic, large-scale generation.
\textbf{SFT} and \textbf{Online RL} show support for finetuning and reinforcement learning.}
\label{tab:framework_comparisons}

\vspace{3pt}

\renewcommand{\arraystretch}{0.97}
\setlength{\tabcolsep}{2.5pt}

\resizebox{\linewidth}{!}{
\begin{tabular}{l r c c c c c c c}
\toprule
\multirow{2}{*}{\textbf{Framework}} &
\multirow{2}{*}{\textbf{\# Tasks}} &
\multicolumn{5}{c}{\textbf{Environments}} &
\multicolumn{2}{c}{\textbf{Training}} \\
\cmidrule(lr){3-7} \cmidrule(lr){8-9}
& &
\textbf{Struct.} &
\textbf{Non-struct.} &
\textbf{POMDP} &
\textbf{Multi} &
\textbf{Scalable} &
\textbf{SFT} &
\textbf{Online} \\
& & \textbf{Obs.} &
\textbf{Obs.} &
& \textbf{Domain} & \textbf{Episodes}
& & \textbf{RL} \\
\midrule
\emph{\textbf{Evaluation-only}} & & & & & & & & \\
\quad OSWorld \citep{xie2024osworld} & 369  & \cmark & ~ & \cmark & ~ & ~ & ~ & ~ \\ 
\quad LIBERO \citep{liu2023libero} & 130  & ~ & \cmark & \cmark & ~ & \cmark & ~ & ~ \\ 
\quad VideoGameBench \citep{zhang2025videogamebench} & 23  & \cmark & \cmark & \cmark & ~ & ~ & ~ & ~ \\ 
\quad LMGame-Bench \citep{hu2025lmgame} & 6  & \cmark & \cmark & \cmark & ~ & ~ & ~ & ~ \\ 
\midrule
\emph{\textbf{Evaluation and Training}} & & & & & & & & \\
\quad VLABench \citep{zhang2025vlabench} & 100 & ~ & \cmark & \cmark & ~ & \cmark & \cmark & \cmark \\ 
\quad VLM-Gym \citep{chen2025g1} & 4  & \cmark & ~ & ~ & ~ & \cmark & \cmark & \cmark \\ 
\quad KORGym \citep{shi2025korgym} & 6  & \cmark & ~ & \cmark & ~ & \cmark & ~ & \cmark \\ 
\quad VisualAgentBench \citep{liu2024visualagentbench} & 5  & \cmark & \cmark & \cmark & \cmark & ~ & \cmark & \cmark \\ 
\quad VAGEN \citep{wang2025vagen} & 5 & \cmark & \cmark & \cmark & \cmark & \cmark & ~ & \cmark \\ 
\midrule
\quad \textbf{\ours{} (Ours)} & 17 & \cmark & \cmark & \cmark & \cmark & \cmark & \cmark & \cmark \\ 
\bottomrule
\end{tabular}
}

\end{table*}

%% file: sec/2_gym.tex
\section{\ours}
\label{sec:vlmgym}

\input{tables/task_overview}
\textbf{\ours} contains $17$ visually interactive environments. 
Each environment exposes initialization parameters that control task configuration and difficulty. 
We provide a high-level overview of the environments in \cref{tab:task_overview} and detailed descriptions with visualizations in \cref{appx:task_details}. 
\ours{} is built on top of the Gymnasium framework \citep{2016gym, towers2024gymnasium}, the same library underlying MuJoCo \citep{todorov2012mujoco} and Atari \citep{bellemare13arcade}. 
Since vision–language agents can interpret images, read instructions, and produce free-form text, we extend Gymnasium with the following enhancements:

\myparagraph{Function-Conditioned Action Space.}
Instead of the discrete or continuous action vectors used in standard Gymnasium environments, we represent actions as function calls with parameters (\textit{e.g.}, \texttt{('swap', (1, 2))}, \texttt{('rotate', (30.5, 20.4, 15.1))}). 
This abstraction allows models to leverage their function-calling capabilities and compose strategies across domains.

\myparagraph{Function Instructions.}
Each task defines a set of functions and their parameter spaces. 
To enable zero-shot rollouts, we provide a natural-language description of these functions and their argument constraints as part of the initial prompt before the model takes its first action. Instructions for each task are shown in \cref{appx:task_details}.

\myparagraph{Environment Feedback.}
In addition to visual transitions, the environment provides textual feedback describing the effect of each action (\textit{e.g.}, ``invalid format," ``out of bounds," ``executed"). 
This helps models with weaker visual perception better ground their actions.

\label{para:solver}
\myparagraph{Solver.}
We implement heuristic multi-step solvers that complete each task using the available actions. 
The solver supports (1) multiple solving strategies and (2) optional stochasticity, enabling the generation of diverse demonstration trajectories for supervised fine-tuning.
See \cref{appx:solver} for the solver design of each task.

\noindent Together, these design choices yield a highly customizable interface. 
Each task can define its own action functions, instruction set, and solver, while the unified \texttt{step} function handles parsing, validation, execution, and feedback (\cref{alg:step} in \cref{appx:interface}). 
This modular structure makes it easy to add new tasks, vary action spaces, and generate visual and textual supervision for VLM agents.

%% file: tables/task_overview.tex
\begingroup
\setcitestyle{numbers}
\begin{table}[t]
\centering
\setlength{\tabcolsep}{3pt}
\caption{
\textbf{\ours{} environments.} For each environment, we specify (1) \textbf{Domain}: whether observations come from \textbf{Real} or \textbf{Synthetic} images, (2) \textbf{Observability (Obs.)}: \textbf{Full} or potentially \textbf{Partial}, (3) \textbf{Dynamics (Dyn.)}: \textbf{Known} vs. \textbf{Unknown} dynamics, (4) \textbf{Parameters (P.)}: number of difficulty parameters, and (5) \textbf{Available Actions}.}
\resizebox{\linewidth}{!}{
\begin{tabular}{llllrl}
\toprule
\textbf{Environment} & \textbf{Domain} & \textbf{Obs.} & \textbf{Dyn.} & \textbf{P.} & \textbf{Available Actions} \\
\midrule

Colorization \scriptsize\citep{zhang2016colorful} & Real & Full & Known & 1 &
\texttt{rotate}$(\theta)$; \texttt{saturate}$(\delta)$; \texttt{stop}() \\

Counting \scriptsize\citep{gupta2019lvis} & Real & Full & Known & 2 &
\texttt{mark}$(x,y)$; \texttt{undo}(); \texttt{guess}$(N)$; \texttt{stop}() \\

Jigsaw \scriptsize\citep{gidaris2018unsupervised} & Real & Full & Known & 2 &
\texttt{swap}$((r_1,c_1),(r_2,c_2))$; \texttt{reorder}$([\dots])$; \texttt{stop}() \\

Matchstick Equation \scriptsize\citep{8924b34498964e6ca74a2661a102b275} & Synthetic & Full & Known & 1 &
\texttt{move}$([i,s,j,t])$; \texttt{undo}(); \texttt{stop}() \\

Matchstick Rotation \scriptsize\citep{10.1145/1054972.1055055} & Synthetic & Full & Unknown & 3 &
\texttt{move}$([dx,dy,d\theta])$; \texttt{stop}() \\

Maze 2D \scriptsize\citep{Koul2024mazeworld} & Synthetic & Full & Known & 2 &
\texttt{move}$(d)$; \texttt{stop}() \\

Maze 3D \scriptsize\citep{Koul2024mazeworld} & Synthetic & Partial & Known & 2 &
\texttt{move}$(0)$; \texttt{turn}$(d)$; \texttt{stop}() \\

Mental Rotation 2D \scriptsize\citep{COOPER197375} & Real & Full & Known & 1 &
\texttt{rotate}$(\theta)$; \texttt{stop}() \\

Mental Rotation 3D \textsc{(Cube)} \scriptsize\citep{ramakrishnan2024does, ShepardMetzler1971} & Synthetic & Partial & Known & 3 &
\texttt{rotate}$([dy,dp,dr])$; \texttt{stop}() \\

Mental Rotation 3D \textsc{(Objaverse)} \scriptsize\citep{ShepardMetzler1971, objaverse} & Synthetic & Partial & Known & 1 &
\texttt{rotate}$([dr,dp,dy])$; \texttt{stop}() \\

MuJoCo Fetch \textsc{(Pick-and-Place)} \scriptsize\citep{todorov2012mujoco} & Synthetic & Partial & Unknown & 0 &
\texttt{move}$([x,y,z])$; \texttt{gripper}$(g)$; \texttt{stop}() \\

MuJoCo Fetch \textsc{(Reach)} \scriptsize\citep{todorov2012mujoco} & Synthetic & Partial & Unknown & 0 &
\texttt{move}$([x,y,z])$; \texttt{stop}() \\

Patch Reassembly \scriptsize\citep{Golomb1994Polyominoes} & Synthetic & Full & Known & 2 &
\texttt{place}$(p,r,c)$; \texttt{remove}$(p)$; \texttt{stop}() \\

Referring Dot-Pointing \scriptsize\citep{kazemzadeh-etal-2014-referitgame} & Real & Full & Known & 0 &
\texttt{mark}$(x,y)$; \texttt{stop}() \\

Sliding Block \scriptsize\citep{Spaans2009} & Synthetic & Full & Known & 1 &
\texttt{move}$(b,d)$; \texttt{stop}() \\

Video Unshuffle \scriptsize\citep{goyal2017something, misra2016shuffle} & Real & Full & Known & 3 &
\texttt{swap}$(i,j)$; \texttt{reorder}$([\dots])$; \texttt{stop}() \\

Zoom-In Puzzle \scriptsize\citep{Baird1970} & Real & Full & Known & 5 &
\texttt{swap}$(i,j)$; \texttt{reorder}$([\dots])$; \texttt{stop}() \\

\bottomrule
\end{tabular}
}
\label{tab:task_overview}
\end{table}
\vspace{-1em}
\endgroup

%% file: sec/3_evaluation.tex
\section{Evaluating Frontier Models with \ours}
\label{sec:eval}

In this section, we evaluate vision-language models on \ours{}. 
We describe our evaluation setup in \cref{subsec:eval_setup} and present results and observational analysis in \cref{subsec:eval_result}.

\subsection{Evaluation Setup}
\label{subsec:eval_setup}

We evaluate $12$ vision-language models spanning three categories:  
\textbf{proprietary} (Gemini~3~Pro~\citep{gemini3}, Gemini~2.5~Pro~\citep{GoogleDeepMind2025_Gemini2_5Pro}, GPT-5~\citep{OpenAI2025_GPT5SystemCard}, Claude~Sonnet~4~\citep{team2025claude}, Grok~4~Fast~\citep{xAI2025_Grok4ModelCard}, Qwen-VL-Max~\citep{bai2025qwen2});  
\textbf{open-weight} models (Qwen3-VL-235B-Instruct~\citep{yang2025qwen3}, GLM-4.5V~\citep{hong2025glm}, Llama-4-Maverick~\citep{touvron2023llama}, Qwen-2.5-VL-72B-Instruct~\citep{bai2025qwen2}, Gemma~3-27B-Instruct~\citep{team2025gemma}); and  
\textbf{specialized} models targeted at GUI/game environments (UI-Tars-1.5-7B~\citep{qin2025ui}).  
We access all proprietary and hosted models through OpenRouter and thus ensure a consistent prompting interface and inference pipeline.
We additionally evaluate models that we finetune on solver demonstrations. We provide details of the supervised finetuning setup in \cref{subsec:sft_exp}.

All models are evaluated in a multi-turn manner.  
At each step $t$, the model receives the full history
\begin{equation}
\label{eq:history}
H_t = \bigl(I,\; \{(o_\tau, a_\tau, f_\tau)\}_{\tau < t}\bigr),
\end{equation}
where $I \in \mathbb{R}^{L_I}$ is the task instruction, $o_\tau \in \mathbb{R}^{H \times W \times C}$ the observation, $a_\tau \in \mathbb{R}^{L_a}$ the action, and $f_\tau \in \mathbb{R}^{L_f}$ the environment feedback.  
The model then outputs an action $a_t$.  
If it outputs the stop action, the environment terminates and returns a binary reward indicating task success.
In addition, we limit the number of interaction steps to 20 for the easy setting and the tasks of Dot-Pointing and Fetch-Reach, 30 for the hard setting and Fetch Pick-n-Place task. 
All tasks are designed to be solvable within these limits, and the environment explicitly provides the number of remaining steps as part of its feedback.
We also ensure that the length of interaction history is within models' context window.
We evaluate each model on 70 episodes per task and setting (\textit{i.e.}, easy, hard).

\subsection{Result and Analysis}
\label{subsec:eval_result}

\myparagraph{Frontier VLMs Fail on \ours{}.}
We show the per-task success rate and the average task success rate of the frontier models in Figure~\ref{fig:main_result} and Figure~\ref{fig:main_result_avg}, respectively.

\input{figs/main_result_average}
\input{figs/main_result}

Even the best-performing frontier model, Gemini-3-Pro, achieves only 46.61\% on \ours{} (Easy) and 26.00\% on \ours{} (Hard), indicating that \ours{} poses a significant challenge for existing models.

\myparagraph{Model Specialization.}
We compare the 3 strongest models\footnote{Gemini 3 Pro is excluded from this detailed comparison, as it was released after this analysis concluded.}: Gemini 2.5 Pro, GPT-5, Qwen3-VL-235B Instruct. 
GPT-5 shows the best ability to handle long-context visual interactions. 
This is reflected in its stronger performance on matchstick rotation where the scale is unknown, its higher scores overall on the hard setting (\cref{fig:main_result_avg}), and its visibly longer tail in the number of steps taken to successfully solve tasks compared to the other models (\cref{fig:eval_steps_density}).
Gemini 2.5 Pro is good at low-level visual perception.
This is reflected in its strongest performance on Jigsaw, Maze 2D, Zoom-In Puzzle, and Sliding Block, all of which demand tight spatial alignment, accurate correspondence of local patterns, and sensitivity to subtle visual cues.
Qwen-3-VL is in particular capable of object localization (\textit{e.g.}, strongest in Referring Dot-Pointing).

Examining the step count distribution (smoothed density curve) for successful trajectories across models (\cref{fig:eval_steps_density}), we found that most models (\textit{i.e.}, Gemini 2.5 Pro, Claude Sonnet 4, and Llama-4-Maverick) only peaked around 3-5 steps, followed by a sharp drop in successful trajectories when they spend more steps.
This indicates limited capability in effectively handling long-context multi-step visual interactions.

\myparagraph{Common Failure Patterns.}
We identify recurring failures using automated failure discovery methods~\citep{dunlap2025vibecheck, stringsight}, which employ a VLM annotator (GPT-4.1) to extract negative behaviors from each trajectory and cluster them into categories observed across datasets. This analysis reveals four failure types that appear consistently across multiple tasks (see \cref{appx:stringsight} for details):

\vspace{-0.5ex}

\noindent\textit{\colorbox{lightyellow}{(1) Restricted action space and action looping:}} models often rely on a single repeated operation or fixed-magnitude action, such as continually moving in the same direction in Fetch Pick \& Place, using ``swap'' in Jigsaw instead of ``reorder'', or rotating by the same angle in Mental Rotation 3D and Match Rotation rather than converging to an optimal magnitude.

\vspace{-0.5ex}

\noindent\textit{\colorbox{lightyellow}{(2) State mismanagement:}} models fail to maintain or update internal state across steps. They ignore textual or environmental feedback, revisit previously explored areas, or repeat illegal actions despite prior errors—for example, continuing to move into a wall after being told they have collided, or repeating invalid moves in the Match Equation, Sliding Block, and Toy Maze 2D tasks.

\vspace{-0.5ex}

\noindent\textit{\colorbox{lightyellow}{(3) Early termination:}} the model terminates before the maximum steps despite not reaching the goal.

\vspace{-0.5ex}

\noindent\textit{\colorbox{lightyellow}{(4) Failure to use visual or spatial information:}} models ignore the visual information provided,  such as the target leaving the frame or the item being successfully aligned (\textit{e.g.}, Mental Rotation).

%% file: figs/main_result_average.tex
\begin{figure*}[t]
\centering
\vspace{-4ex}
\begin{minipage}[t]{0.60\textwidth}
    \centering
    \includegraphics[width=1.0\linewidth, trim={0mm 0 0 0}, clip]{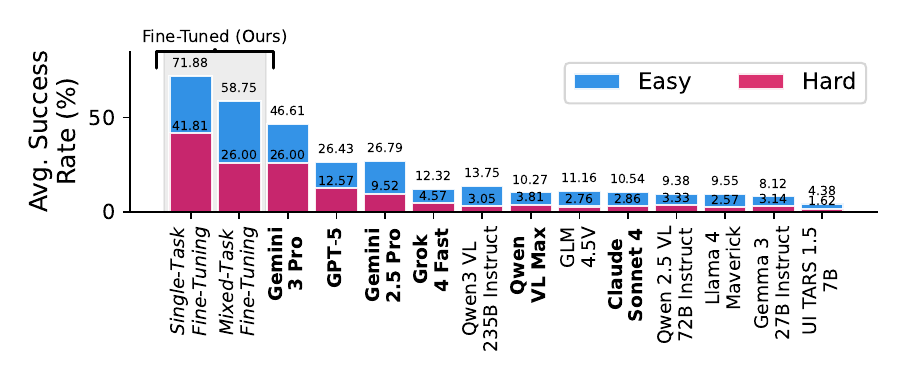}
    \vspace{-4ex}
    \caption{\textbf{Average task success rate for frontier models and our finetuned models}. Proprietary models are in \textbf{bold} and our finetuned models are \textit{italicized}.}
    \label{fig:main_result_avg}
\end{minipage}
\hfill
\begin{minipage}[t]{0.38\textwidth}
    \centering
    \includegraphics[width=\linewidth]{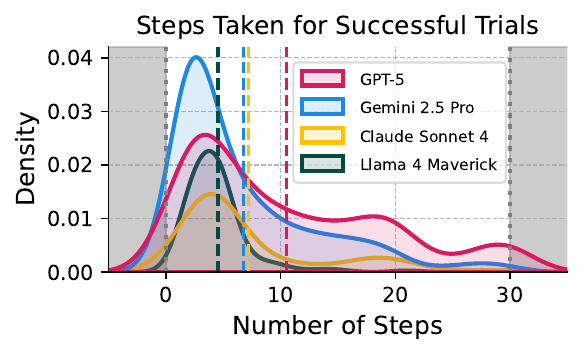}
    \vspace{-4ex}
    \caption{\textbf{Density curve of steps taken for successful trajectories.} Colored dashed line marks each model’s mean number of steps.}
    \label{fig:eval_steps_density}
\end{minipage}
\vspace{-1em}
\end{figure*}

%% file: figs/main_result.tex
\begin{figure*}[h]
\includegraphics[width=1.0\linewidth, trim={15mm 0 0 0}]{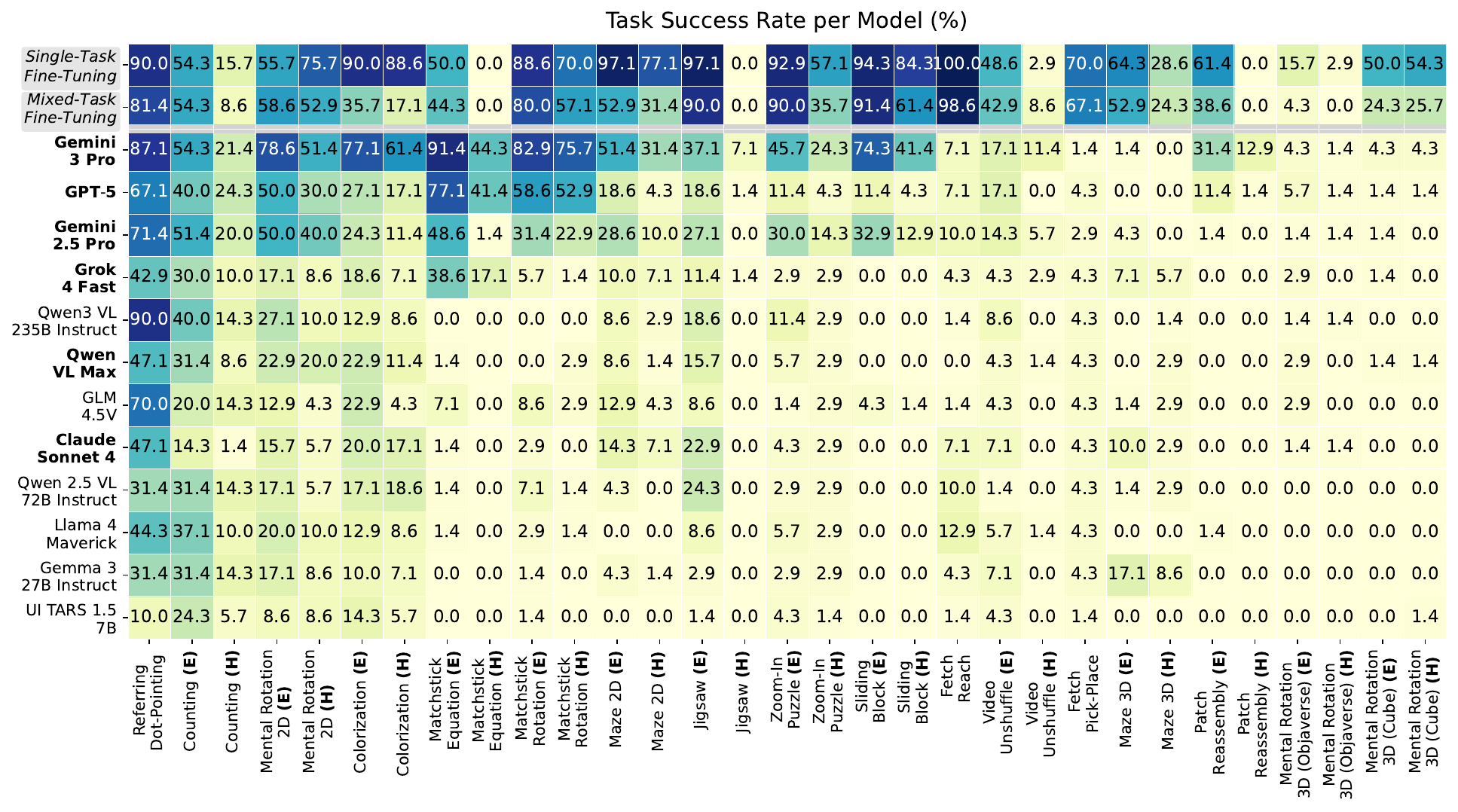}
\vspace{-2ex}
    \caption{\textbf{Task success rate of frontier and finetuned models}. Proprietary models are shown in \textbf{bold}, and our finetuned models in \textit{italics}. \textit{(E)} and \textit{(H)} denote easy and hard task settings. Darker cells indicate higher success rates. Models are ordered by average task performance (top = better), and tasks by average model performance, excluding our finetuned ones (right = harder). \colin{task average across models and model average across tasks as a row and a column. just keep the percentage, no decimal places}}
    \label{fig:main_result}
\vspace{-4ex}
\end{figure*}

%% file: sec/4_diagnosis.tex
\section{Diagnosing Frontier Models with \ours{}}
\label{sec:diagnosis}

In this section, we show the flexibility of \ours{} by presenting controlled diagnoses of how different designs of multi-step interactions can drastically change frontier models' performance and provide our conjectures on why these designs make a difference.

We perform diagnoses with the two best performing proprietary models, \textit{i.e.}, GPT-5 \citep{OpenAI2025_GPT5SystemCard} and Gemini 2.5 Pro \citep{GoogleDeepMind2025_Gemini2_5Pro}, and the two best performing open-weight models, \textit{i.e.}, Qwen3-VL-235B Instruct \citep{yang2025qwen3} and GLM-4.5V \citep{hong2025glm}.

\subsection{Turns to Keep in Conversation History}
\label{subsec:turn_to_keep}
Vision--language models are known to degrade with long visual context \citep{wang2025multimodal, wu2024visual, sharma-etal-2024-losing}. 
This creates a dilemma: while long histories provide more information about the environment (\textit{e.g.}, 3D layouts, unknown dynamics), they also introduce redundant observations that may harm performance.
We study this trade-off in Maze2D, Sliding Block, MuJoCo Fetch Reach, and Matchstick Rotation, where history provides useful signals such as textual feedback (\textit{e.g.}, invalid actions) or correspondence between action magnitude and perceptual effect, but also introduces stale information.

\input{figs/diag_history}

As shown in \cref{fig:diag_history}, models benefit from including a limited number of previous turns up to roughly four, following a drop when given the full unbounded history. 
This indicates that expanding visual context helps multi-step visual decision-making only to a point, after which irrelevant or stale observations become detrimental.
We also observe task-specific idiosyncrasies: 
Gemini~2.5~Pro scales well in Maze2D, 
GPT-5 scales well on Matchstick Rotation, while Sliding Block exhibits clear \emph{reverse scaling} for Gemini~2.5~Pro. 
These highlight that the value of interaction history is both task-dependent. \jiaxin{todo: add insights from stringsight here. ? what are u doing junyi grammer error, too mu ohch}

\subsection{Representing Observation in Text}
\label{subsec:text_repr}
\jz{worth putting a visualization here, it's not obvious right now for the reader} \jiaxin{In the appendix or main paper?} \jz{main paper} \colin{i don't think we have space}
Inspired by prior work examining how different task representations affect agent performance \citep{hu2025lmgame, shi2025korgym, ruoss2024lmact}, we select four symbolic tasks--Matchstick Equation, Maze~2D, Patch Reassembly, and Sliding Block--and implement alternate versions rendered entirely in ASCII (sample ASCII visualizations are provided in \cref{appx:ascii_task_visualizations}).  
This allows tasks to be solved without any visual encoding module.   

\input{figs/diag_text}

The results in \cref{fig:diag_text} show that GPT-5 substantially improves in most tasks, often achieving $3$ -- $4\times$ higher success rates than in the visual setting, suggesting that its main bottleneck lies in visual grounding rather than long-horizon reasoning.
Gemini~2.5~Pro shows mixed behavior: two tasks do not exhibit significant performance change, one task improves, and one task degrades, indicating possible limitations in both perception and planning. 
Open-weight models struggle across all tasks in both modalities, indicating general weaknesses in long-horizon decision-making regardless of representation.
Interestingly, Matchstick Equation exhibits a \emph{reverse} trend: all models perform substantially better with the visual representation than with ASCII, likely because the figlet-style ASCII has irregular shapes and spacing that create distorted glyphs which models are known to struggle with \citep{stojanovski2025reasoning}.

\subsection{Removal of Text-based Feedback}
\label{subsec:feedback}
Humans can infer action consequences directly from visual changes \citep{michotte1963perception}, but it remains unclear whether VLMs can do the same.
To study this, we select four tasks---Maze~3D, Maze~2D, Sliding Block, and Matchstick Equation---in which the environment feedback $f$ (see \cref{eq:history}) provides not only formatting errors but also constraint violations (\textit{e.g.}, hitting a wall in Maze, sliding a block into an occupied cell).  
We remove this textual feedback and evaluate model using only visual state transitions; results are shown in \cref{fig:diag_feedback}.

\input{figs/diag_feedback}

All models show consistent drops in average performance.
This indicates that models struggle to infer action validity directly from visual transitions.  
These findings show that current VLMs depend heavily on text-based feedback during visually interactive decision-making and are less sensitive to pure visual feedback.

\subsection{Providing Final Goal at Beginning}
\label{subsec:final_goal}
Providing the solution image upfront simplifies the tasks to visually aligning current observations with a known target, shifting the difficulty from reasoning to visual perception and tool-calling.
We test this on five tasks, Patch Reassembly, Jigsaw, Colorization, Zoom-In Puzzle, and Matchstick Equation, where constructing the goal observation involves significant effort.
For these tasks, we augment the instruction with the ground-truth final observation $o_{gt}$, and show results in \cref{fig:diag_goal}.

\input{figs/diag_goal}

Across tasks, models improve substantially, indicating that a major bottleneck lies in \textit{constructing or imagining the target state}.
However, performance remains far from perfect, indicating additional limitations beyond reasoning, such as fine-grained visual perception and action calling.
Surprisingly, GPT-5 and Gemini~2.5~Pro \textit{underperform} on the Zoom-In Puzzle and Matchstick Equation when the final goal observation is provided, often terminating early despite visible misalignment.
A follow-up test confirms this stems from visual misjudgment due to limited visual perception: we queried Gemini 2.5 Pro on 100 pairs of initial and final-goal observations with the prompt ``\textit{Do the two images look exactly the same?}''
and it incorrectly judged images as identical $80\%$ and $57\%$ of the time for these tasks, versus only $18\%$, $2\%$, and $0\%$ for Colorization, Jigsaw, and Patch Reassembly.
This confirms that perception errors can \textit{invert} the expected benefit of an explicit goal observation.

%% file: figs/diag_history.tex
\begin{figure*}[h]
\centering
\includegraphics[width=\linewidth]{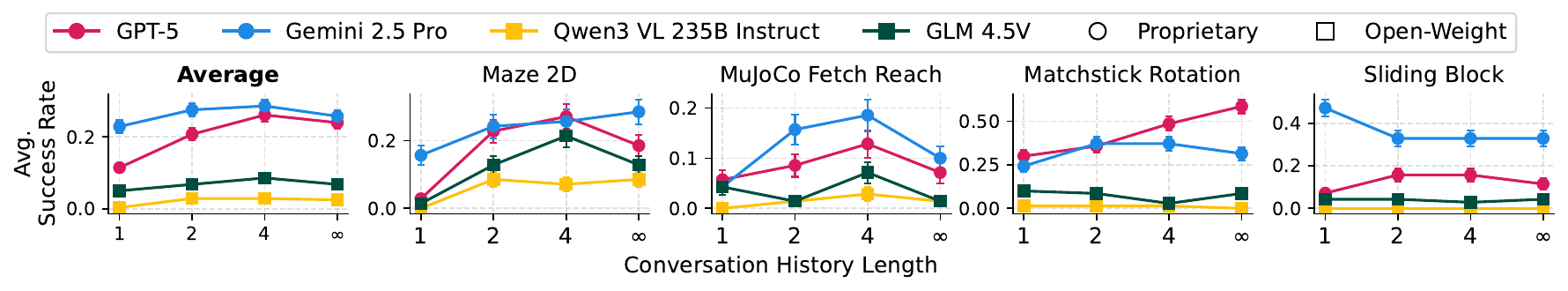}
\vspace{-5ex}
\caption{\textbf{Effect of truncating conversational context on model performance}. The settings $1$, $2$, $4$, and $\infty$ correspond to retaining only the current turn, the current $+$ previous turn, the current $+$ previous $3$ turns, and the full history, respectively. Error bars show the standard error of the mean.}
\label{fig:diag_history}
\vspace{-2ex}
\end{figure*}

%% file: figs/diag_text.tex
\begin{figure*}[h]
\includegraphics[width=\linewidth]{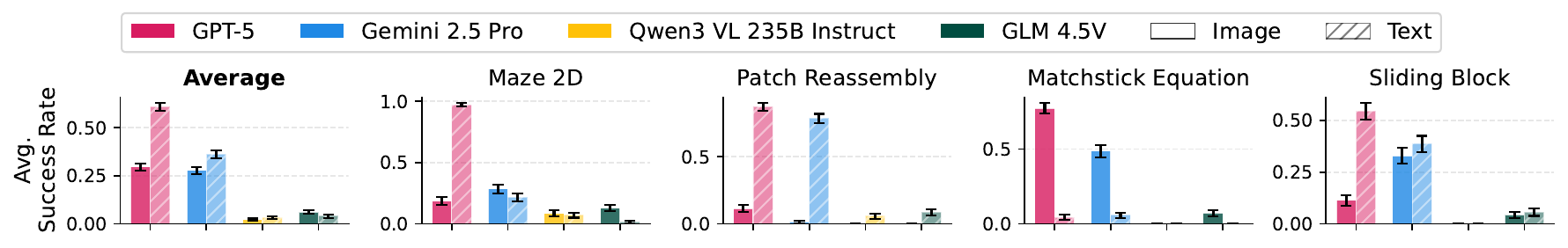}
\vspace{-5ex}
\caption{\textbf{Effect of visualizing observations with ASCII (\emph{text})}. “Image’’ and “Text’’ denote the observation modalities. Error bars show the standard error of the mean.}
\vspace{-3ex}
\label{fig:diag_text}
\end{figure*}

%% file: figs/diag_feedback.tex
\begin{figure*}[h]
\includegraphics[width=\linewidth]{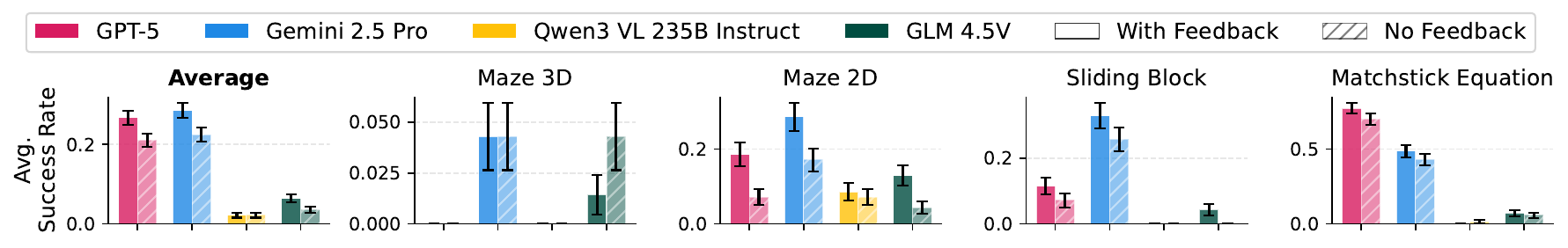}
\vspace{-5ex}
    \caption{\textbf{Effect of removing text-based environment feedback}. “With Feedback’’ includes environment feedback describing action execution at each turn; “No Feedback’’ removes this channel. Error bars show the standard error of the mean.}
    \label{fig:diag_feedback}
\vspace{-2ex}
\end{figure*}

%% file: figs/diag_goal.tex
\begin{figure*}[ht]
\includegraphics[width=\linewidth]{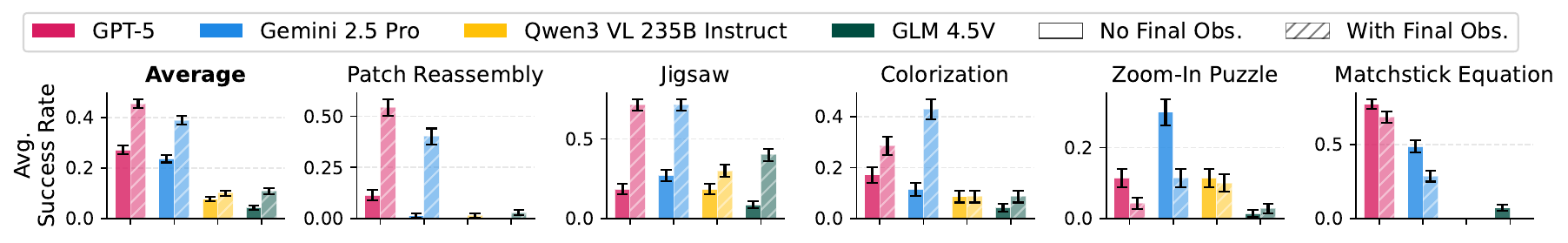}
\vspace{-5ex}
\caption{\textbf{Effect of providing the final goal observation at the beginning of the episode}. “No Final Obs.’’ and “With Final Obs.’’ denote settings without and with access to the goal observation at the start. Error bars show the standard error of the mean.}
\label{fig:diag_goal}
\vspace{-2ex}
\end{figure*}

%% file: sec/5_training.tex
\section{Training with \ours}
\label{sec:train}

We describe our supervised fine-tuning experiments with \ours{}, present results, and provide insights on generalization, module specificity, and data curation.

\subsection{Supervised Fine-Tuning Experiments}
\label{subsec:sft_exp}

\myparagraph{Setup.}
We generate demonstration trajectories for supervised fine-tuning using the multi-step solver described in \cref{para:solver}.
We apply two preprocessing filters: (1) discarding trajectories that fail to complete the task, and (2) removing trajectories with initial states overlapping the test split to prevent data leakage.

We evaluate two fine-tuning configurations: \textit{single-task} and \textit{mixed-task}.
In the \textit{single-task} setting, we fine-tune a separate model for each task, whereas in the \textit{mixed-task} setting, a single model is trained jointly on all tasks.
Notably, demonstrations are sourced exclusively from the easy difficulty level; thus, performance on the hard setting serves as a metric for difficulty generalization.
All experiments employ Qwen2.5-VL-7B-Instruct~\citep{bai2025qwen2} with full-parameter fine-tuning, a global batch size of $64$, a learning rate of $1 \times 10^{-5}$, and \texttt{bf16} precision.
Models are trained for $1,500$ steps in the single-task setting and $5,000$ steps in the mixed-task setting.
We utilize LlamaFactory~\citep{zheng2024llamafactory} for all data preprocessing and training orchestration.

\myparagraph{Results.} 
As shown in \cref{fig:main_result,fig:main_result_avg}, finetuned models achieve state-of-the-art performance on most tasks, validating both the learnability of our environments and the effectiveness of our multi-step solvers.  
These gains confirm that current VLMs can substantially benefit from structured, solver-generated demonstrations in visually grounded multi-step settings.

\subsection{Stronger Base Model Generalizes Better}
\label{subsec:base_generalization}
\input{figs/train_variant_generalization}
Existing work has discussed the limitations of supervised finetuning \citep{ross2011reduction} and found that it exhibits limited generalization to task variants \citep{caccia2024finetuning, deng2023mind2web, jang2022bc}.  
This motivates re-examining generalization in the context of modern VLMs, whose capabilities may shift the boundary of what supervised finetuning can or cannot retain.

To this end, we select a set of environments where the easy-to-hard difficulty gap introduces substantial state changes (\textit{e.g.}, more views in the Zoom-In Puzzle, more patches in Patch Reassembly, larger maze sizes; details in \cref{appx:task_variant_generalization}).  
We finetune Qwen2.5-VL-7B-Instruct and Qwen3-VL-8B-Instruct on the same mixed-task training data using identical optimization hyperparameters (see \cref{subsec:sft_exp}), and report performance in \cref{fig:train_variant_gen}.

As shown, both models achieve comparable performance on the easy variants they were trained on (\textit{e.g.}, $0.59$ vs.\ $0.64$), but the more recent Qwen3-VL generalizes substantially better to the harder variants, nearly doubling the success rate on average relative to Qwen2.5-VL.
This trend highlights that newer VLMs provide stronger out-of-distribution generalization in multi-step visual decision-making despite being finetuned on an identical setup.

\subsection{Vision and LLM Both Matter}
\input{figs/train_modules}
Classic perception–action theories emphasize that fine-grained visual encoding and temporal integration are jointly necessary for interactive behavior \citep{Gibson1979TheEA}.  
We examine whether this holds for VLMs by fine-tuning variants that modify either the vision encoder or the LLM backbone to isolate each module’s contribution, where the vision encoder provides fine-grained perceptual features and the LLM performs temporal integration across steps.  

As shown in \cref{fig:train_modules}, most tasks benefit from fine-tuning both components, with the LLM contributing the larger performance gain—particularly in tasks with partial observability or unknown environment dynamics.  
This highlights that temporal reasoning and history integration remain the primary bottlenecks for current VLMs, while strong fine-grained visual encoding is necessary (\textit{e.g.,} Zoom-In Puzzle primarily benefits from vision finetuning) but often not sufficient for multi-step decision-making.

\subsection{Importance of Information-Revealing Behaviors for SFT Curation}

\input{figs/train_curation}

Not all experiences contribute equally to decision-making: trajectories that reveal hidden states or disambiguate perceptual aliasing are often far more valuable \citep{NIPS1994_d2ed45a5, fujii-etal-1998-selective}.
We ask whether inducing such information-revealing behaviors during supervised finetuning helps VLMs form more accurate state representations.
We evaluate this on two tasks, Matchstick Rotation (unknown dynamics) and Mental Rotation~3D Objaverse (partial observability), with results in \cref{fig:train_curation_matchstick_rotation,fig:train_curation_mental_rotation}.

In Matchstick Rotation, the baseline demonstrations perform three stochastic moves toward the target.
In contrast, the information-revealing demonstrations first perform two unit-scale steps to expose the correspondence between action magnitude and perceptual effect before executing the final aligning move.
This structured exploration raises success from $32.9\%$ to $70.0\%$.

In Mental Rotation, the baseline trajectories rotate along each principal axis once to reach the goal, while the information-revealing ones deliberately fully rotate along each axis to expose the full 3D geometry before settling on the target orientation.
This strategy improves performance in both metrics.
To verify that gains are not simply due to longer trajectories, we further continue training on baseline demonstrations starting from the model already finetuned with information-revealing data.
Performance deteriorates in this setting, confirming that the observed improvements stem from the \emph{informative structure} of the demonstrations rather than quantity or length.
These results highlight that SFT effectiveness depends on whether demonstrations induce state-disambiguating behaviors, not merely on the number of examples.

%% file: figs/train_variant_generalization.tex
\setlength{\columnsep}{6pt}
\setlength{\intextsep}{0pt}

\begin{wrapfigure}{r}{0.5\linewidth}
\vspace{-1.5ex}   %
\centering
\includegraphics[width=\linewidth]{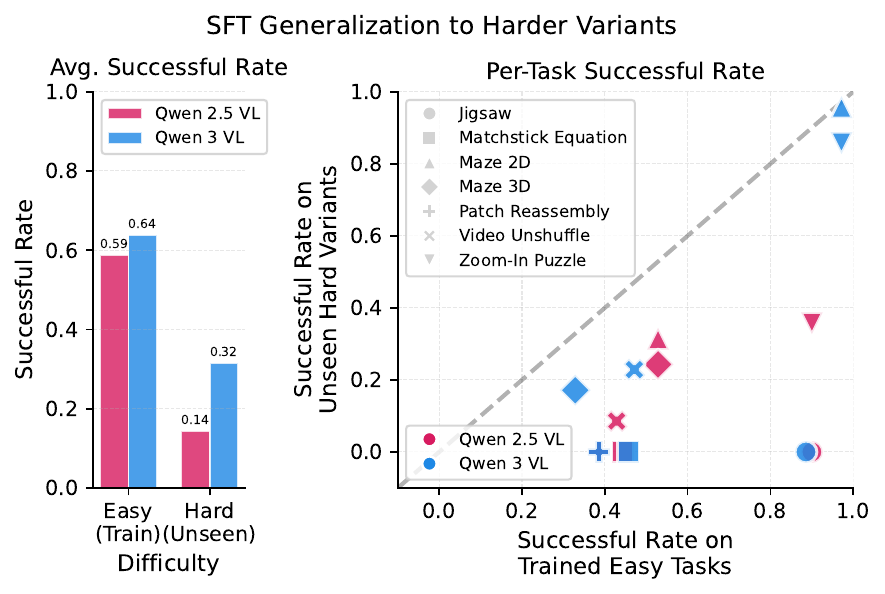}
\vspace{-5ex}   %
\caption{\textbf{Generalization to unseen difficulty from mixed-task supervised finetuning.}
\emph{(Left)}: average success rate across 7 tasks in easy (seen) and hard (unseen) settings for Qwen2.5-VL and Qwen3-VL.
\emph{(Right)}: task-level plot comparing success rates; X-axis = easy, Y-axis = hard.}
\label{fig:train_variant_gen}
\vspace{-0ex}   %
\end{wrapfigure}

%% file: figs/train_modules.tex
\setlength{\intextsep}{0pt}

\begin{wrapfigure}{r}{0.5\linewidth}
\vspace{-8ex}   %
\centering
\includegraphics[width=\linewidth]{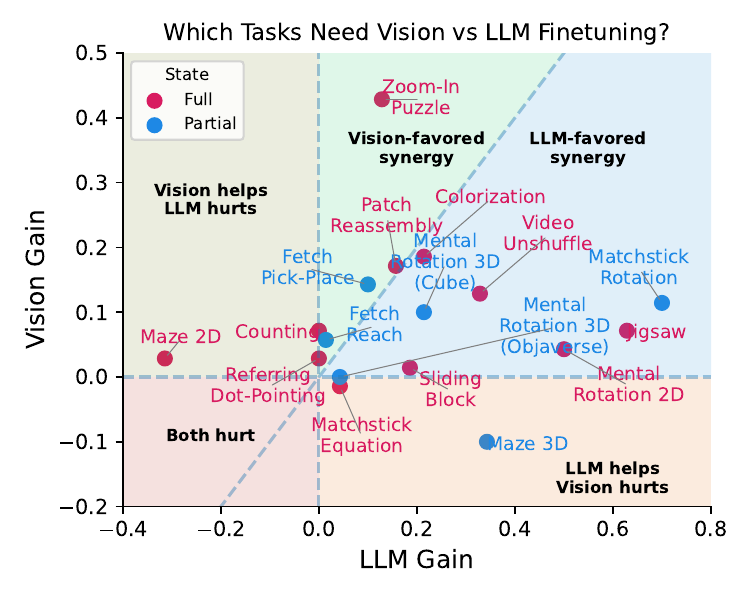}
\vspace{-5ex}   %
\caption{\textbf{Tasks benefiting from finetuning different modules.}
“Vision Gain’’ and “LLM Gain’’ denote improvements from jointly finetuning both components, compared to finetuning only the \textit{LLM} or the \textit{vision} part. The dashed line ($y=x$) divides vision-favored (above) and LLM-favored (below) synergy. ``Full’’ and ``Partial’’ denote whether observability and dynamics are fully known.}
\label{fig:train_modules}
\vspace{-1.5ex}   %
\end{wrapfigure}

%% file: figs/train_curation.tex
\begin{wrapfigure}{r}{0.5\linewidth}
\begin{minipage}{\linewidth}
    \vspace{-1ex}

    \includegraphics[width=\linewidth]{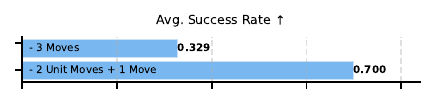}
    \vspace{-4ex}
    \captionof{figure}{\textbf{Effect of data curation strategies on task performance when 
    \colorbox{lightyellow}{environment dynamics are unknown.}}
    Numbers represent average task success (higher is better).
    “3 Moves’’ and “2 Unit Moves + 1 Move’’ are two curation strategies.}
    \label{fig:train_curation_matchstick_rotation}

    \vspace{-0ex}

    \includegraphics[width=\linewidth]{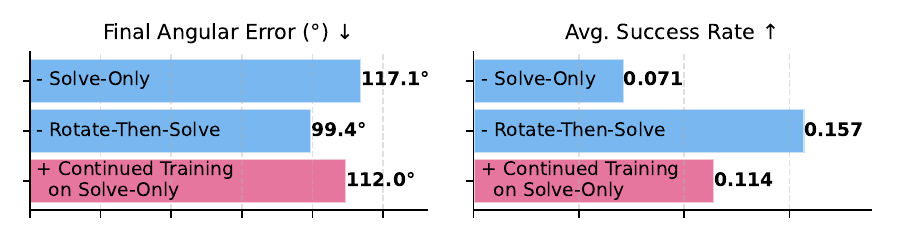}
    \vspace{-4ex}
    \captionof{figure}{\textbf{Effect of data curation strategies on task performance when 
    \colorbox{lightyellow}{environment is partially observable.}}
    “Solve-Only’’ and “Rotate-Then-Solve’’ are two curation strategies, and
    “Continued Training on Solve-Only’’ denotes further finetuning on Solve-Only after training on Rotate-Then-Solve.
    \emph{(Left)}: final angular error on the test set (lower is better).
    \emph{(Right)}: average task success rate (higher is better).}
    \label{fig:train_curation_mental_rotation}

\end{minipage}
\vspace{-2ex}
\end{wrapfigure}

%% file: sec/6_related_work.tex
\section{Related Work}

\colin{might need a refocus -- haven't checked}
\jz{shorten this section to a single paragraph, and refer to the appendix for the full version}
The development of foundation models \citep{achiam2023gpt,team2023gemini,team2025claude,dubey2024llama,yang2025qwen3}, particularly vision-language models (VLMs) \citep{hong2025glm,bai2025qwen2,team2025kimi,zhu2025internvl3,team2025gemma,liu2023visual,alayrac2022flamingo,li2022blip} and vision-language-action models (VLAs) \citep{black2410pi0, bjorck2025gr00t, team2025gemini, kim2024openvla, octo_2023, huang2025otter, lbmtri2025, zhou2025vision, niu2024llarva, shi2025diversity}, has reshaped how AI agents perceive, make decisions, and act across physical and simulated environments. To properly assess the capabilities and limitations of the models, a plethora of benchmarks have been developed.

Early benchmarks such as Atari, OpenAI Gym, and DeepMind Lab \citep{mnih2013playing,2016gym,towers2024gymnasium,beattie2016deepmind} were developed to evaluate vision-based control and decision-making in fully observable environments. These platforms laid the groundwork for reinforcement learning but focused primarily on low-level motor control. Subsequent efforts extended these ideas to robotic manipulation and navigation, introducing partially observable, multi-task, and long-horizon settings that better reflect real-world complexity \citep{yu2020meta, ahmed2020causalworld, shridhar2020alfred, li2024behavior, cao2025learn, liu2023libero, khanna2024goat, choi2024lota, yang2025embodiedbench, ehsani2021manipulathor, szot2021habitat, srivastava2022behavior, mees2022calvin, mandlekar2021matters, james2020rlbench}. These modern suites enable training and evaluation of multi-task imitation learning and meta-learning policies across diverse embodiment and task horizons.

Concurrently, many VLM benchmarks have been developed to probe models’ cognitive and perceptual limits. Early efforts focused on visual question answering—first as multiple-choice tasks and later as open-ended reasoning \citep{yue2024mmmu, yue2025mmmu, liu2024mmbench, li2024seed, chen2024we}. As visual grounding and reasoning improved, newer benchmarks began representing actions through text instead of fixed, predefined action spaces, enabling studies of the interplay between perception, reasoning, and control \citep{wang2025jigsaw, jang2024videowebarena, liuagentbench, shi2025korgym, stojanovski2025reasoning, abdulhai2023lmrl, coelho2025deepresearchgym}. For example, G1 \citep{chen2025g1} introduces VLM-Gym, a suite of visual game environments with unified interfaces and adjustable difficulty. Broader evaluation suites such as VisualAgentBench \citep{liu2024visualagentbench}, EmbodiedBench \citep{yang2025embodiedbench}, and WebArena \citep{zhou2023webarena} aggregate tasks across embodied control, graphical interfaces, and visual reasoning, challenging agents with multi-step planning and tool use.

\ours{} unifies reasoning and control under an RL-style “gym” paradigm, combining 17 multimodal tasks spanning visual puzzles, spatial reasoning, manipulation, and grounding. Each environment includes an oracle solution to ensure solvability and allow synthetic trajectory generation for post-training. Moreover, \ours{} introduces controllable difficulty along with targeted diagnostics—such as history utilization, representation variants, feedback specificity, and perception–action causality—allowing researchers to examine not only whether models fail but also what causes failures. We hope these designs enable more systematic analysis of VLMs and VLAs across domains and levels of interactivity (\cref{tab:framework_comparisons}).

%% file: sec/7_conclusion.tex
\vspace{-0.1em}
\section{Conclusion}
\vspace{-0.2em}
We present \ours{}, a unified suite of $17$ visually interactive environments that challenge and train vision–language models in multi-step visual decision-making. 
\ours{} establishes a rigorous playground for building the next generation of multimodal agents, bridging perception and reasoning toward more capable, adaptive visual intelligence.

\section*{Acknowledgement}
We thank Trevor Darrell, Jacob Steinhardt, Yichuan Wang, Ryan Yixiang Wang, Haiwen Feng, Baifeng Shi, Jihan Yang and Ziqiao Ma for their feedback. We also thank OpenRouter for its support in our model evaluation. Authors, as part of their affiliation with UC Berkeley, were supported by gifts from Accenture, AMD, Anyscale, Broadcom, Cisco, Google, IBM, Intel, Intesa Sanpaolo, Lambda, Lightspeed, Mibura, Microsoft, NVIDIA, Qualcomm, Samsung SDS, and SAP. Authors, as part of their affiliation with UC Berkeley, were supported in part by the National Science Foundation, US Department of Defense, and/or the Berkeley Artificial Intelligence Research (BAIR) industrial alliance program. This research was also developed with funding from the Defense Advanced Research Projects Agency (DARPA) under Contract No. W912CG-24-C-0011. The views, opinions and/or findings expressed are those of the authors and should not be interpreted as representing the official views or policies of any sponsor, the Department of Defense, or the U.S. Government.

%% file: sec/8_appendix.tex
\appendix
\clearpage
\etocdepthtag.toc{mtappendix}
\etocsettagdepth{mtchapter}{none}    %
\etocsettagdepth{mtappendix}{section} %

\begingroup
  \parskip0pt
  \etocsettocstyle{\section*{Table of Contents}}{} %
  \tableofcontents
\endgroup

\makeatletter
\setlength{\@fptop}{0pt}               %
\setlength{\@fpsep}{8pt}               %
\setlength{\@fpbot}{0pt plus 1fil}     %

\setlength{\@dblfptop}{0pt}            %
\setlength{\@dblfpsep}{4pt}            %
\setlength{\@dblfpbot}{0pt plus 1fil}  %
\makeatother

\input{appendices/solver_design}

\input{appendices/task_details}

\input{appendices/ascii}

\input{appendices/interface_better}

\clearpage
\section{Analyzing Model Failures}
\label{appx:stringsight}

\noindent
We run StringSight~\cite{dunlap2025vibecheck, stringsight}, a pipeline for automatically uncovering failure cases and comparing models. It uses a VLM annotator (GPT-4.1) to extract behaviors from each trace (e.g., ``uses \texttt{move(1,1)} for all 20 steps'') and clusters these behaviors into higher-level patterns (e.g., ``repeats the same action''). Examples of discovered cluster descriptions are shown in Table~\ref{tab:stringsight_discovery_clusters}. We then manually examine the top failure cases for each task and identify four common failure modes across all tasks.

\input{tables/stringsight}
\noindent\textit{\colorbox{lightyellow}{(1) Restricted action space and action looping:}} models often rely on a single repeated operation or fixed-magnitude action, such as continually moving in the same direction in Fetch Pick \& Place, using ``swap'' in Jigsaw instead of ``reorder'', or rotating by the same angle in Mental Rotation 3D and Match Rotation rather than converging to an optimal magnitude.

\noindent\textit{\colorbox{lightyellow}{(2) State mismanagement:}} models fail to maintain or update internal state across steps. They ignore textual or environmental feedback, revisit previously explored areas, or repeat illegal actions despite prior errors—for example, continuing to move into a wall after being told they have collided, or repeating invalid moves in the Match Equation, Sliding Block, and Toy Maze 2D tasks.

\noindent\textit{\colorbox{lightyellow}{(3) Early termination:}} the model terminates the episode before the maximum steps, despite not reaching the goal.

\noindent\textit{\colorbox{lightyellow}{(4) Failure to use visual or spatial information:}} models ignore the visual information provided,  such as the target leaving the frame or the item being successfully aligned (\textit{e.g.}, Mental Rotation). 

Finally, we quantify the prevalence of each failure mode by having a VLM annotator (GPT-4.1) label each trace for these behaviors (a trace may exhibit multiple behaviors).

\vspace{-1em}

\begin{table*}
\begin{tcolorbox}[title=System Prompt: Labeling failure modes in traces, colback=white, colframe=black]

You are an expert model behavior analyst. Your task is to meticulously analyze the trace of a large language model to identify whether it contains any of the following behaviors:

\vspace{0.5em}

\begin{itemize*}

    \item \textbf{Restricted action space and action looping}: The model keeps repeating the same or nearly identical action without making progress. Look for consecutive turns with the same command or sequence of commands, movements by the same amount, or the same tool being used even when it is ineffective.\par\vspace{0.8em}\\

    \item \textbf{State mismanagement}: The model forgets or ignores what it already learned in earlier steps. It may revisit old states, contradict past reasoning, or repeat mistakes it was corrected for (e.g. being told it hit a wall and then continuing to move forward in the same direction). Do not include if this is simply action looping, where the model is repeating the same action without making progress; this is specifically when the model is ignoring feedback or not adjusting its behavior based on its previous actions, but is still issuing different commands.\par\vspace{0.8em}\\

    \item \textbf{Early termination}: The model stops too early. Early means terminating before the maximum number of steps is reached.\par\vspace{0.8em}\\

    \item \textbf{Failure to use visual or spatial information}: The model ignores visible or spatial cues. This applies to traces where either an image or ASCII art is provided, and the model does not react to changes in the scene. For example, if the object leaves the frame, but the model continues to move towards it. Look for actions that contradict what's visually or spatially clear. Do not include if the model is simply action looping, where the model is repeating the same action without making progress; this is specifically when the model is not utilizing the visual information when it is available. If the trace does not provide visual information (either images or ASCII), do not include this label.\par\\

\end{itemize*}

If the trace contains any of the behaviors, return a list of objects with the following structure. If a trace has more than one behavior, return a list of objects with the structure below for each behavior. It the trace contains none of the behaviors, return an empty list.

\textbf{JSON Output Structure}
\begin{verbatim}
[
  {
    "property_description": which behavior is present in the trace,
    "reason": an explanation of the exact behaviors in the trace
              that fall under the property_description (1-2 sentences),
    "evidence": "What exactly in the trace exhibits this property?
                 Include quotes/tool calls/actions when possible."
  }
]
\end{verbatim}

\end{tcolorbox}
\vspace{-1.5em}
\end{table*}

\paragraph{Frequency of failures.} 

\begin{wrapfigure}{r}{0.4\linewidth}
    \vspace{-2ex}
    \centering
    \includegraphics[width=\linewidth, trim=0cm 0cm 0cm 0cm, clip]{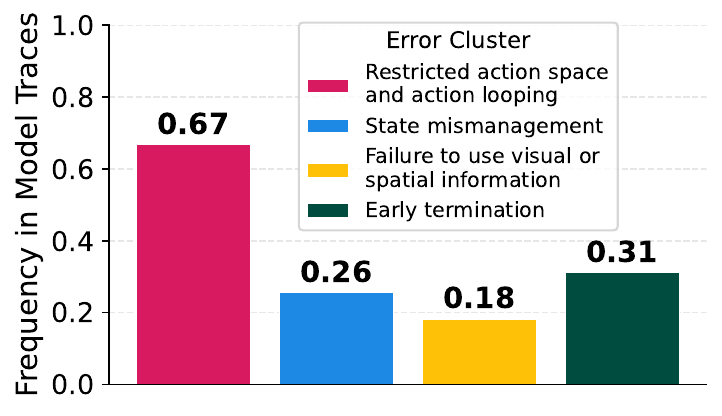}
    \vspace{-2ex}
    \caption{\textbf{Frequency of failure patterns.}}
    \label{fig:freq_failure}
\end{wrapfigure}
\begin{figure*}[h]
\begin{center}

    \begin{minipage}[t]{0.42\textwidth}
        \centering
        \includegraphics[width=\linewidth]{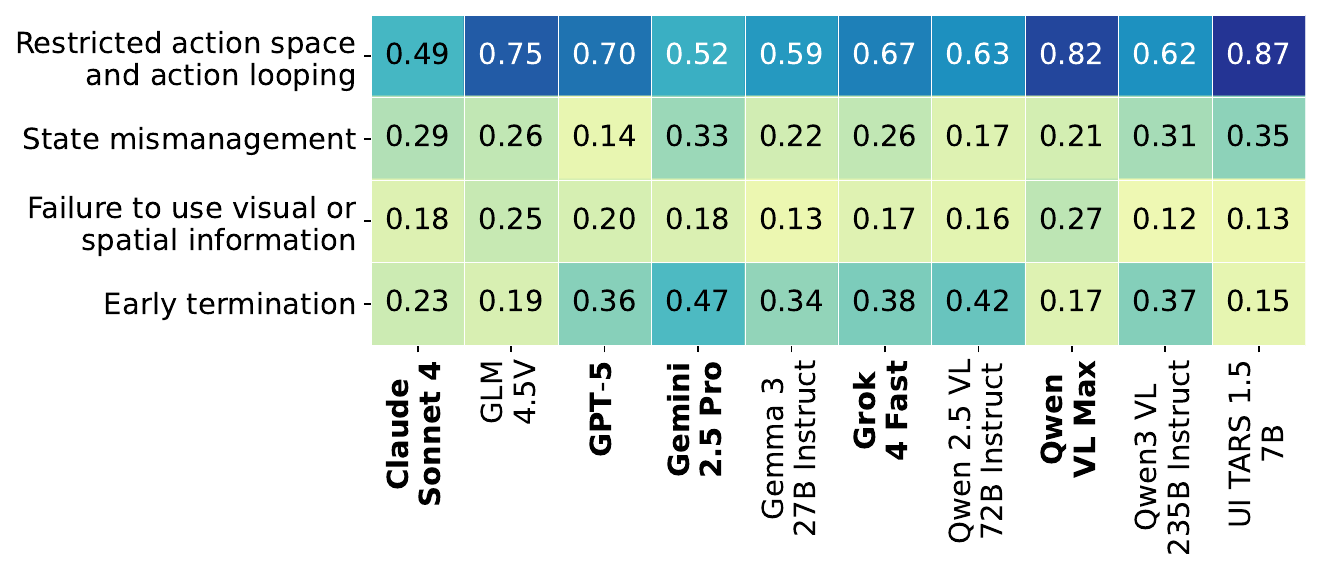}
        \vspace{0.5ex}
        \textbf{(a)} Frequency of failure patterns per model.
        \label{fig:freq_failure_model}
    \end{minipage}
    \hfill
    \begin{minipage}[t]{0.57\textwidth}
        \centering
        \includegraphics[width=\linewidth]{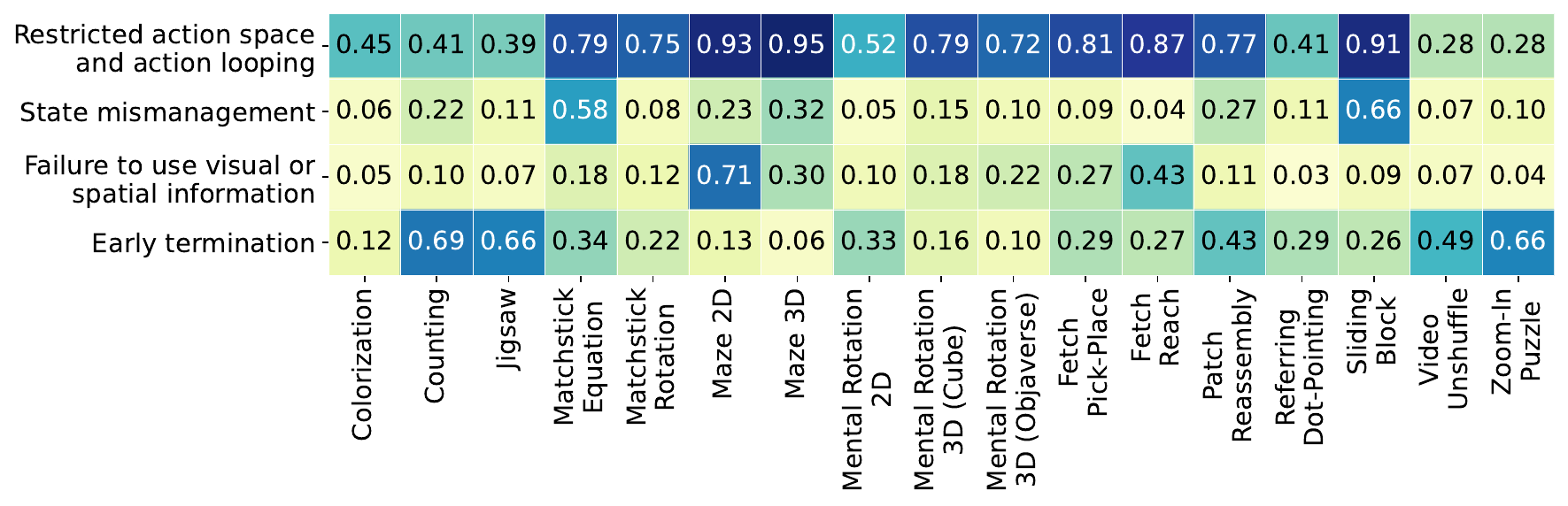}
        \vspace{0.5ex}
        \textbf{(b)} Frequency of failure patterns per task on easy variants.
        \label{fig:freq_failure_task}
    \end{minipage}

    \vspace{-1ex}
    \captionsetup[figure]{hypcap=false}
    \captionof{figure}{\textbf{Detailed Analysis of Failure Patterns by Model and Task}.}
    \label{fig:combined_model_task_analysis}
\end{center}
\end{figure*}

Figure~\ref{fig:freq_failure} shows the proportion of traces that contain each failure. We see that action looping is very common, occurring in more than 60\% of traces, followed in frequency by early termination, state mismanagement, and failure to use visual or spatial information. Looking at how the frequency of the behaviors changes compared across tasks, we see in Figure~\ref{fig:combined_model_task_analysis} (b) that certain tasks, like Matchstick Equation and Sliding Block, result in a particularly large amount of action repetition and state mismanagement failures, likely due to the difficulty of the task and the frequency of invalid moves. We additionally see that tasks like the Maze task, which provide clear visual signals of task progress, have a very high (up to 70\%) rate of ignoring this important visual information and high action repetition. Based on this information, we see that often when a model is uncertain, it defaults to repeating its previous moves, regardless of the visual or language feedback it is given from previous turns. This is further supported in Figure~\ref{fig:combined_model_task_analysis} (a), which shows that weaker models like UI TARS 1.5 7B have very high rates of action looping (87\%) and state mismanagement (35\%).

\vspace{0.5em}

We additionally find interesting cases of early termination, such as giving up on the task entirely, where the model says things like ``I give up.'' and ``I'm stopping. This is unsolvable''. 
These specific instances of giving up happen much more often for hard tasks like Matchstick Equation, indicating that the models’ limited task comprehension leads them to question whether a solution exists in the current instance.
We also see this phenomenon occur more often in Gemini and Gemma models, which we suspect is because these models are chattier and more anthropomorphic and thus may express their internal reasoning more often than others.

\subsection{Failure changes per ablation}

\begin{figure*}[t]
    \centering

    \begin{subfigure}[b]{0.24\linewidth}
        \centering
        \includegraphics[height=3.7cm]{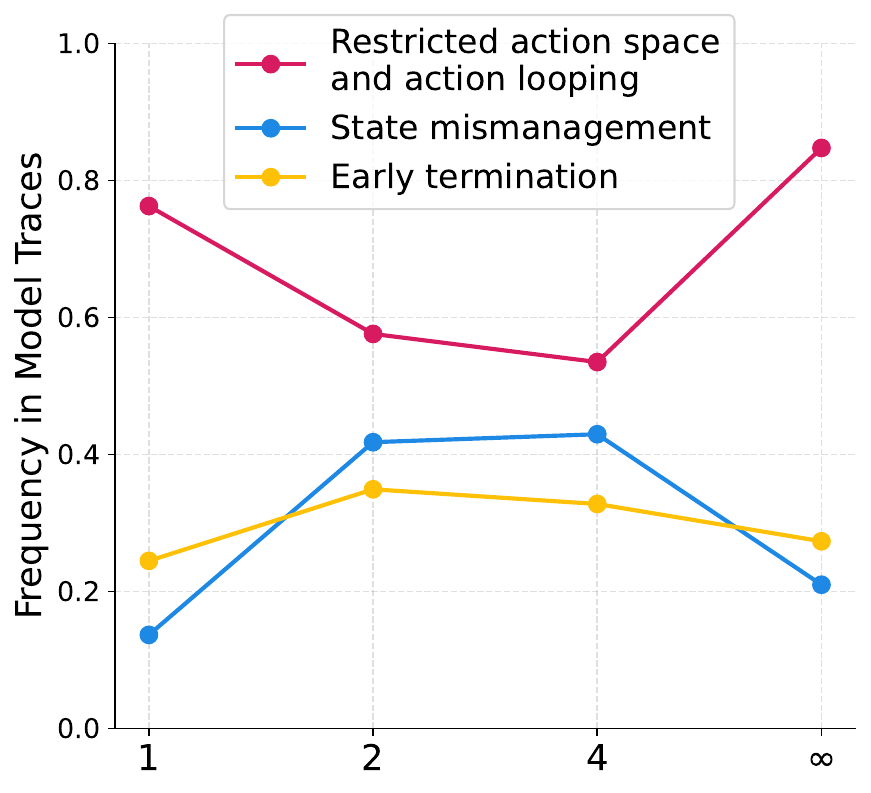}
        \caption{Different amounts of chat history.}
        \label{fig:freq_failure_history}
    \end{subfigure}
    \hfill
    \begin{subfigure}[b]{0.2\linewidth}
        \centering
        \includegraphics[height=3.7cm]{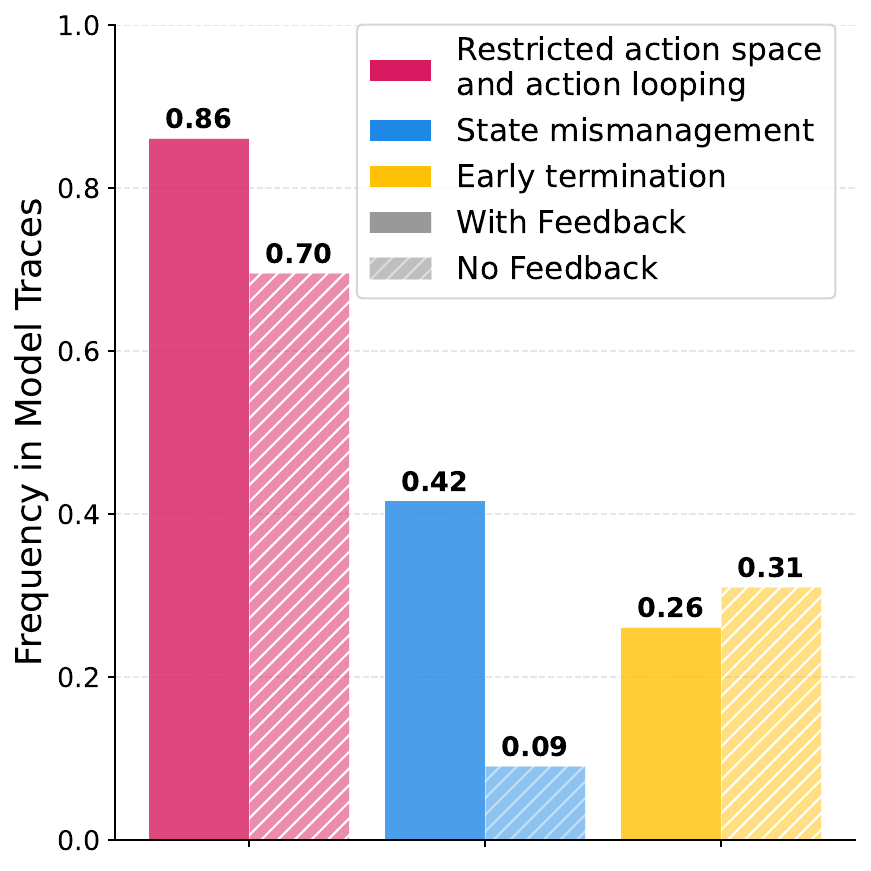}
        \caption{Feedback vs. no feedback.}
        \label{fig:freq_failure_feedback}
    \end{subfigure}
    \hfill
    \begin{subfigure}[b]{0.24\linewidth}
        \centering
        \includegraphics[height=3.7cm]{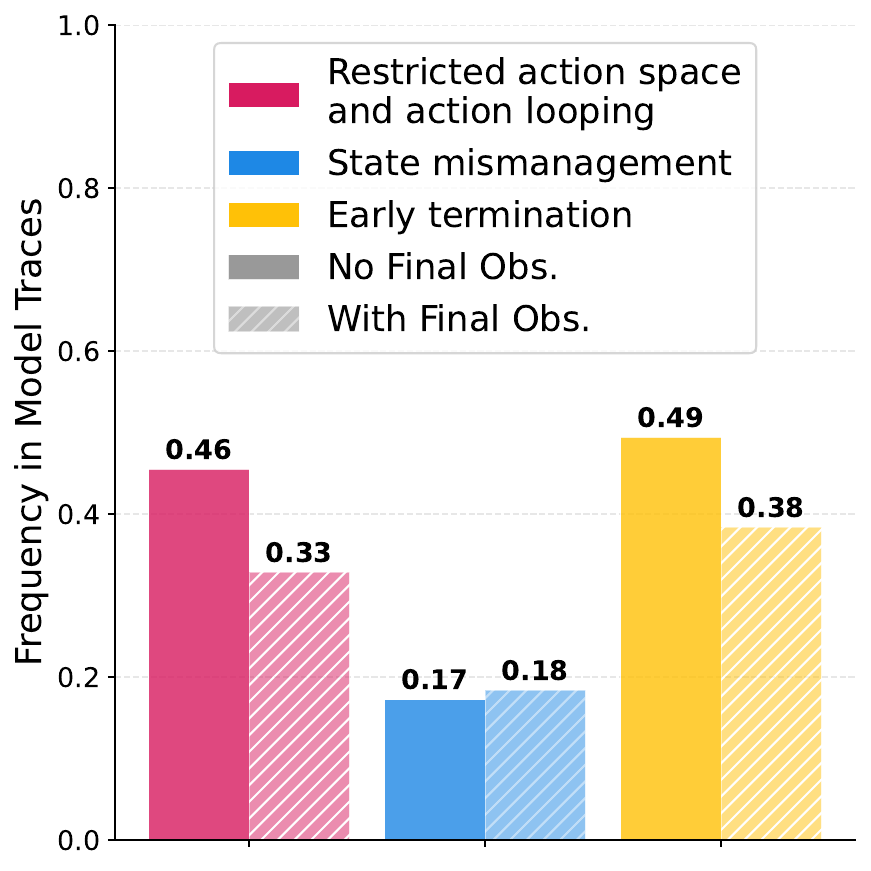}
        \caption{Ground-truth obs. at the start.}
        \label{fig:freq_failure_gt}
    \end{subfigure}
    \begin{subfigure}[b]{0.24\linewidth}
        \centering
        \includegraphics[height=3.7cm]{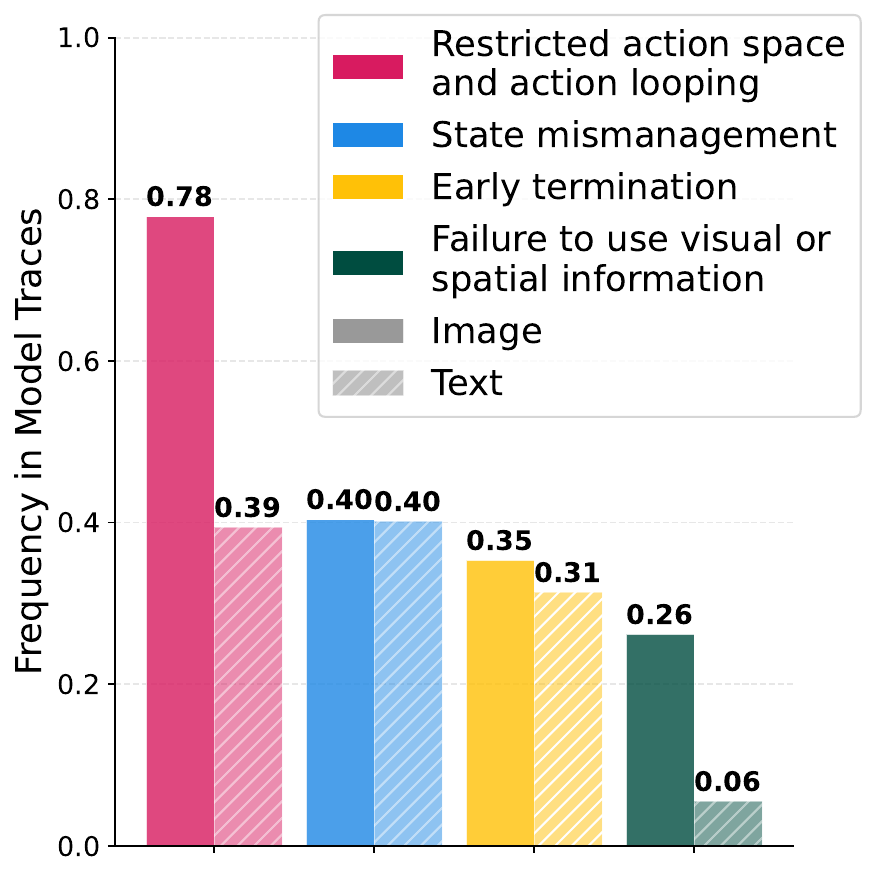}
        \caption{Text vs. image representation.}
        \label{fig:freq_failure_text}
    \end{subfigure}
\vspace{-0.5em}
    \caption{\textbf{Failure-pattern frequency under different information settings.}
             Due to cost, images were only analyzed in the original split, thus the ``failure to use visual or spatial information'' case is removed in all but the text vs image representation ablation (d).}
    \label{fig:freq_failure_grid}
    \vspace{-0.5em}
\end{figure*}

To examine the effects of our ablations on model behavior, we run the failure labeling pipeline above on the ablations described in Section~\ref{sec:diagnosis} and show the comparison in Figure~\ref{fig:freq_failure_grid}. We find the following:

\textit{Different amounts of chat history (Figure~\ref{fig:freq_failure_history}):}
As more history is given, the model is less likely to repeat immediate actions, but still suffers from state mismanagement. We suspect the decreased action looping occurs because the model has a default action (e.g., moving left), so with no history, it continues to repeat this move. With history, it is less likely to immediately repeat prior actions, but after a certain amount of context, the model struggles to manage earlier state and reverts to its default behavior. This is reflected by action looping decreasing when full history is given, consistent with decreased performance under full history.

\textit{Feedback vs. no feedback (Figure~\ref{fig:freq_failure_feedback}):}
When no feedback is provided, the model is less likely to terminate. Inspecting these traces shows that this is largely due to a reduction in “giving up,” since the model often gives up when told its moves are invalid. We also observe decreases in action looping and state mismanagement, which is surprising given that overall performance decreases without feedback. This suggests the presence of additional failure modes not captured by our taxonomy, which we leave for future work.

\textit{Ground truth state given at the beginning (Figure~\ref{fig:freq_failure_gt}):}
When given the ground truth state at the start of the task, the model is less likely to “guess’’ or give up early, reflected in lower rates of action looping and early termination.

\textit{Image vs. text representation (Figure~\ref{fig:freq_failure_text}):}
For tasks aside from Matchstick Equation, models process visual information more effectively when it is presented as text rather than an image. The large reduction in action looping suggests that this text-based representation provides clearer guidance for selecting actions. \\

\subsection{Failure Trajectories Visualization}
\label{sec:stringsight_traj}
Using StringSight, we visualize the trajectory for each failure type. In each trajectory, we show the prompt, the image, the models' raw output, and the action parsed from the raw output. We also show the output from StringSight for each trajectory, tagged ``Reason'' and ``Evidence'' at the top, where ``Reason'' stands for StringSight's reason for classifying this trajectory into a specific failure category, and ``Evidence'' stands for the evidence in the trajectory that leads to the conclusion.

\vspace{0.2em}
\smallskip
\noindent\textit{\colorbox{lightyellow}{(1) Restricted action space and action looping:}} As in \cref{subsubsec:traj_action_looping}, we show a case of action looping of GPT-5 on the Jigsaw task. The model repeatedly takes the same action ``(``swap'', (0, 0), (0, 1))'', resulting in looping behaviors without making any progress.

\smallskip
\noindent\textit{\colorbox{lightyellow}{(2) State mismanagement:}} 
    As in \cref{subsubsec:traj_state}, we show a case of Claude Sonnet 4 on Maze 2D. In Observation 7, the model takes action ``(``move'', 2)'', which leads to the environment feedback ``Cannot move into a wall.'' However, at Observation 16, the model is in the exact same state, disregards the previous feedback, and takes the same action ``(``move'', 2)'' again.

\smallskip
\noindent\textit{\colorbox{lightyellow}{(3) Early termination:}} 
     We show in \cref{subsubsec:traj_early} a case of Gemma 3 27B Instruct on Matchstick Equation, where the model decides to give up and terminate at step 13, while the model is allowed to take 30 steps in total. 

\smallskip
\noindent\textit{\colorbox{lightyellow}{(4) Failure to use visual or spatial information:}} 
    As in \cref{subsubsec:traj_visual}, we show a case of Gemini 2.5 Pro on Mental Rotation 3D (Cube). In the last three steps, after rotating in the wrong direction, the model does not take the visual information into account and continues to rotate in the same direction, which moves the object even farther away from the target position.

\section{Additional Performance Analysis}
\label{appx:analysis}

\myparagraph{Difficulty of Each Task.}
\input{figs/tast_difficulty_ranking}
In~\cref{fig:task_difficulty_ranking}, we compute the average accuracy across models for each task and sort tasks from easiest to hardest based on these averages. 
In general, we found that Referring Dot-Pointing and Counting are the easiest, with models achieving over 20\% accuracy on average, whereas Mental Rotation 3D (Cube), Patch Reassembly, and
Mental Rotation 3D (Objaverse) are the hardest with an accuracy around 1\%. 
Sliding Block, Maze 3D, Fetch Pick-Place, and Video Unshuffle also pose significant challenges for the models, with less than 5\% on average.
This suggests that tasks requiring memory and long-horizon planning, or strong 3D spatial understanding, remain the most difficult for current models.

\myparagraph{Number of Steps.}
In~\cref{fig:num_steps}, we calculate the number of steps taken on all trajectories for each model and calculate the number of correct trajectories (green) and the number of incorrect trajectories (red). There is a clear cutoff on steps 20 (maximum steps allowed for Easy setting) and 30 (maximum steps allowed for Hard setting), indicating that all models tend to reach the maximum number of steps. We also observed a ``U-shaped" trend over the steps for all models, where they tend to either terminate early or continue until the final step.
\input{figs/num_steps}

\input{figs/easy_vs_hard}
\myparagraph{Easy to Hard Performance Drop.} In~\cref{fig:easy_hard}, we calculate the average accuracy in Easy and Hard, respectively, on all models, and then visualize the performance gap between easy and hard on each task.
The biggest Easy to Hard performance drops occur on Counting and Jigsaw.
For Counting, accuracy drops sharply as the number of objects increases.
For Jigsaw, performance drops to near zero as the puzzle changes from 2x2 to 3x3, suggesting that this task can be further scaled to even more difficult n×n configurations.
For some tasks (\textit{e.g.}, Patch Reassembly, Sliding Block, Video Unshuffle), the absolute gap is smaller, likely because Easy performance is already very low ($\approx$0).
These tasks are also naturally scalable in terms of difficulty. For example, increasing the number of patches for Patch Reassembly, the number of blocks for Sliding Block, or the number of frames for Video Unshuffle. 
As VLMs improve and begin to reliably solve the Easy settings, we expect to see larger easy-to-hard gaps on these tasks, and that our gym can be correspondingly scaled to provide harder task variants.

\begin{figure*}[t]
\includegraphics[width=0.955\linewidth]{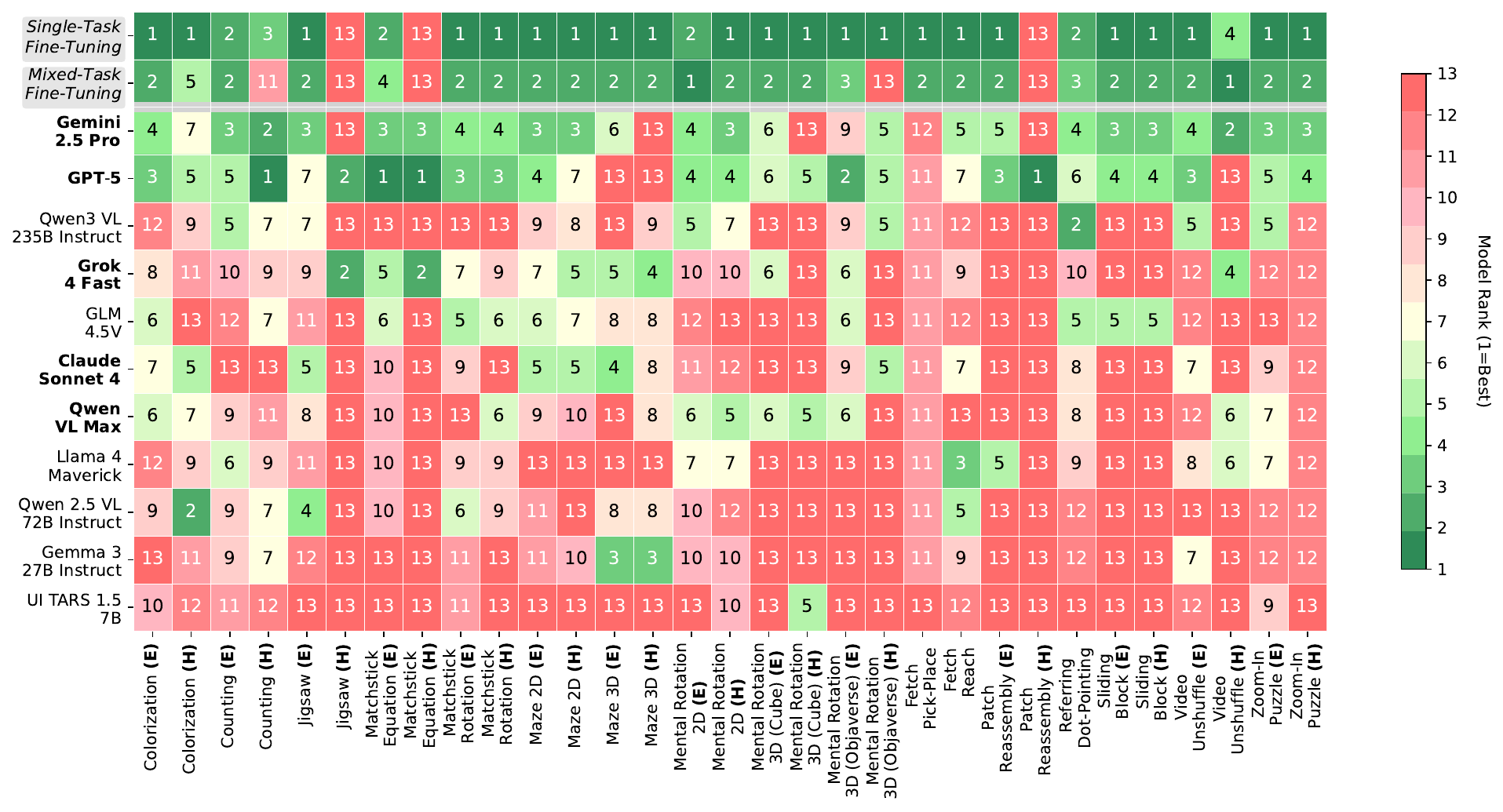}
\vspace{-1ex}
    \caption{\textbf{Model Rankings Per Task.} We rank all the models on each task and show the ranking in the table.}
    \label{fig:ranking}
\end{figure*}
\myparagraph{Model Rankings.}
In~\cref{fig:ranking}, we plot each model’s ranking on every task. When multiple models tie on a task, we assign them the worse (higher) numerical rank. For example, if two models are tied for the best score, we label both as rank 2, so that ties do not overstate how clearly a model is separated from others.
While these mostly align with the global leaderboard, they also reveal clear task-specific strengths and weaknesses. For example, on specialization, Qwen2.5 VL 72B Instruct performs well on Counting (rank 2), Gemma 3 27B Instruct performs well on Maze 3D (rank 3), and Qwen3 VL 235B Instruct performs well on Referring Dot-Pointing. On weakness, despite strong average performance, GPT-5 performs poorly on Video Unshuffle and Maze 3D, while Gemini 2.5 Pro performs poorly on Jigsaw (Hard) and Patch Reassembly.

\clearpage
\setlength{\EnvStepCaseNeedspace}{0.0\textheight}
\setcounter{section}{6}
\EnvRolloutCase[winered]{Sample trajectory for ``Action Looping" (GPT-5)\label{subsubsec:traj_action_looping}}{cvpr2026/figures/stringsight_visualizations/traj_0}{8}
\clearpage

\clearpage
\setlength{\EnvStepCaseNeedspace}{0.2\textheight}
\EnvRolloutCase[winered]{Sample trajectory for ``State Mismanagement" (Claude Sonnet 4)\label{subsubsec:traj_state}}{cvpr2026/figures/stringsight_visualizations/traj_2}{21}
\clearpage

\clearpage
\setlength{\EnvStepCaseNeedspace}{0.3\textheight}
\EnvRolloutCase[winered]{Sample trajectory for ``Early Termination" (Gemma3 27B)\label{subsubsec:traj_early}}{cvpr2026/figures/stringsight_visualizations/traj_3}{14}
\clearpage

\clearpage
\setlength{\EnvStepCaseNeedspace}{0.1\textheight}
\EnvRolloutCase[winered]{Sample trajectory for ``Visual/Spatial Perception" (Gemini 2.5 Pro)\label{subsubsec:traj_visual}}{cvpr2026/figures/stringsight_visualizations/traj_1}{15}
\clearpage

\clearpage
\setcounter{section}{2}
\setcounter{subsection}{0}
\EnvRollout{Colorization\label{subset:colorization}}{cvpr2026/figures/appendix/colorization__ColorizationEnv-v2}{7}
\clearpage

\clearpage
\EnvRollout{Counting\label{subsec:counting}}{cvpr2026/figures/appendix/counting__LVISCountingEnv-v0}{8}
\clearpage

\EnvRollout{Jigsaw\label{subsec:jigsaw}}{cvpr2026/figures/appendix/Jigsaw-v0}{9}
\clearpage

\EnvRollout{Matchstick Equation\label{subsec:matchstick_equation}}{cvpr2026/figures/appendix/match_equation__MatchEquation-v0}{6}
\clearpage

\EnvRollout{Matchstick Rotation\label{subsec:matchstick_rotation}}{cvpr2026/figures/appendix/match_move__MatchRotation-v0}{5}
\clearpage

\EnvRollout{Maze 2D\label{subsec:maze2d}}{cvpr2026/figures/appendix/ToyMaze2DEnv-v0}{18}
\clearpage

\EnvRollout{Maze 3D\label{subsec:maze3d}}{cvpr2026/figures/appendix/ToyMaze3DEnv-v0}{21}
\clearpage

\EnvRollout{Mental Rotation 2D\label{subsec:mr2d}}{cvpr2026/figures/appendix/mental_rotation__mental_rotation-v0}{3}
\clearpage

\EnvRollout{Mental Rotation 3D \textsc{(Cube)}\label{subsec:mr3d_cube}}{cvpr2026/figures/appendix/mental_rotation_3d__mental_rotation-3d}{3}
\clearpage

\EnvRollout{Mental Rotation 3D \textsc{(Objaverse)}\label{subsec:mr3d_objaverse}}{cvpr2026/figures/appendix/mental_rotation-3d_objaverse-v0}{3}
\clearpage

\EnvRollout{MuJoCo Fetch \textsc{(Pick-and-Place)}\label{subsec:mujoco_pick_place}}{cvpr2026/figures/appendix/FetchPickAndPlaceDiscrete-v4}{28}
\clearpage

\EnvRollout{MuJoCo Fetch \textsc{(Reach)}\label{subsec:mujoco_reach}}{cvpr2026/figures/appendix/FetchReachDiscrete-v4}{12}
\clearpage

\EnvRollout{Patch Reassembly\label{subsec:patch_reassembly}}{cvpr2026/figures/appendix/patch_reassembly__PatchReassemblyEnv-v0}{8}
\clearpage

\EnvRollout{Referring Dot-Pointing\label{subsec:referring_dot}}{cvpr2026/figures/appendix/ref_dot__RefCOCOPlusDotEnv-v0}{5}
\clearpage

\EnvRollout{Sliding Block\label{subsec:sliding_block}}{cvpr2026/figures/appendix/sliding_block__SlidingBlockEnv-v0}{14}
\clearpage

\EnvRollout{Video Unshuffle\label{subsec:video_unshuffle}}{cvpr2026/figures/appendix/video_unshuffle__VideoUnshuffleEnv-v0}{4}
\clearpage

\EnvRollout{Zoom-In Puzzle\label{subsec:zoom_in_puzzle}}{cvpr2026/figures/appendix/zoom_in__ZoomInEnv-v0}{5}
\clearpage

%% file: appendices/solver_design.tex
\clearpage
\section{Solver Design}
\label{appx:solver}

This section provides detailed descriptions of the multi-step solvers introduced in \cref{sec:vlmgym} and used for supervised finetuning across all environments.

\myparagraph{Colorization.} The solver computes how far the current hue and saturation are from the target, breaks those differences into small incremental steps, and outputs a sequence of rotate and saturate actions that move steadily toward the correct color. If a target number of steps is requested or if the color is already close enough, it fills the sequence with reversible rotate/saturate pairs that cancel out and don’t change the final state.

\myparagraph{Counting.}
\textit{mark\_all strategy:} The solver places a dot at the center of each target instance, then submits the correct total count and stops. \textit{guess\_only strategy:} The solver directly submits the correct total count and stops, without placing any dots.

\myparagraph{Jigsaw.}
\textit{reorder strategy:} The solver computes a single permutation payload that, when applied via the reorder' action, instantly rearranges the current pieces into their correct target positions. \textit{swap strategy:} The solver generates a minimal sequence of swap' actions by repeatedly finding a misplaced piece and swapping it with the piece at its correct target location. If a target number of steps is requested, it pads this sequence with reversible pairs of swaps (\textit{e.g.}, swapping two pieces and then immediately swapping them back) until the desired length is reached.

\myparagraph{Matchstick Equation.}
\textit{bfs strategy:} The solver finds the shortest possible sequence of move' actions to correct the equation using a Breadth-First Search (BFS) and then stops.
\textit{dfs strategy:} The solver finds a solution using a Depth-First Search (DFS), producing a sequence of move' actions and undo' actions that represent its full exploratory and backtracking process before stopping.
\textit{sos strategy:} The solver first finds the shortest solution path (via BFS), then pads this path by inserting random, reversible detours. Before an optimal step, it takes one or more random move' actions and immediately undo'es them, returning to the optimal path before proceeding.

\myparagraph{Matchstick Rotation.}
The solver first performs one or more translation-only move' actions, which are typically unit-length moves in the general direction of the target. It then executes a final move' action that applies the entire required rotation and corrects any remaining translation error, before stopping.

\myparagraph{Maze 2D.}
The solver uses a graph search algorithm to find the optimal coordinate path from the agent to the target, which is converted into the shortest sequence of move' actions. If a target number of steps is requested, the solver pads this optimal sequence by inserting random, reversible move' pairs (\textit{e.g.}, move up' followed by move down') at valid locations along the path until the desired length is met, before stopping.

\myparagraph{Maze 3D.}
The solver uses a graph search algorithm to find the optimal coordinate path from the agent's location to the target. It then converts this path into the shortest sequence of turn' (left, right, or around) and move' actions required to follow that path, accounting for the agent's current orientation. If a target number of steps is requested, the solver pads this optimal sequence by inserting random, reversible turn' pairs (\textit{e.g.}, turn left' followed by turn right') at locations along the path until the desired length is met, before stopping.

\myparagraph{Mental Rotation 2D.}
The solver first calculates the shortest total rotation angle required to align the current image with the target. If the requested number of steps is 1, it outputs a single rotate' action for that total angle. If a larger number of steps is requested, it stochastically divides the total rotation into that many smaller rotate' actions, which are executed sequentially and sum to the correct total angle, before stopping.

\myparagraph{Mental Rotation 3D \textsc{(Cube)}.}
The solver decomposes the total required rotation into its yaw, pitch, and roll components. It then corrects each component sequentially. Before applying the corrective `rotate' action for a specific axis (\textit{e.g.}, yaw), it first executes a padding sequence of four 90-degree rotations around that same axis. After this 360-degree padding, it applies the single action to correct the yaw. It repeats this pad-then-correct process for the pitch and roll axes, then stops.

\myparagraph{Mental Rotation 3D \textsc{(Objaverse)}.}
The same as Mental Rotation 3D \textsc{(Cube)}.

\myparagraph{MuJoCo Fetch \textsc{(Pick-and-Place)}.}
The solver is a state-machine-based oracle. It follows a sequence: (0) move the gripper to a safe height above the object, (1) open the gripper, (2) descend to the object, (3) close the gripper to grasp. (4) Once grasped, it moves the object directly toward the 3D goal position using a greedy, per-axis strategy (correcting the axis with the largest error at each step). (7) Finally, it holds the object at the target location and stops.

\myparagraph{MuJoCo Fetch \textsc{(Reach)}.}
The solver is a greedy, per-axis oracle. At each step, it identifies the single axis (x, y, or z) with the largest error between the gripper and the goal. It then outputs a `move' action along that single axis to reduce the error, repeating this process until the goal is reached, at which point it stops.

\myparagraph{Patch Reassembly.}
The solver uses a backtracking search to find the optimal sequence of `place' actions that perfectly tile the grid. If a target number of steps is requested, it pads this sequence by repeatedly inserting ``mistake-and-correct" actions: it finds a correct `place' action in the solution, finds a valid wrong location for that piece, and inserts this ``mistake" `place' action immediately before the ``correct" `place' action. If no valid mistakes can be found, it falls back to inserting a `remove' and a duplicate `place' action. This repeats until the desired number of `place' actions is met.

\myparagraph{Referring Dot-Pointing.}
The solver first samples a random pixel from within the target object's segmentation mask and also calculates the mask's center of mass. It then generates a sequence of `mark' actions by linearly interpolating from the random starting point to the center of mass over the requested number of steps. The final action in this sequence places a mark at the exact center of mass, which is then followed by a `stop' action.

\myparagraph{Sliding Block.}
The solver uses a Breadth-First Search (BFS) to find the shortest sequence of `move' actions from the current board state to the target configuration. If a target number of steps is requested, it pads this optimal path by first reconstructing all intermediate board states. At each state, it identifies all valid ``back-and-forth" moves (\textit{e.g.}, move block 1 right, then move block 1 left). It then randomly samples from these opportunities and inserts the required number of `move' and `reverse-move' pairs into the solution path until the desired length is met, before stopping.

\myparagraph{Video Unshuffle.}
\textit{reorder strategy:} The solver computes a single permutation payload that, when applied via the `reorder' action, instantly rearranges the shuffled frames into their correct chronological order, then stops.
\textit{swap strategy:} The solver generates a minimal sequence of `swap' actions to sort the frames. It iterates through the positions, and if a frame is in the wrong place, it finds the correct frame and swaps it into its target position, repeating until all frames are sorted, then stops.

\myparagraph{Zoom-In Puzzle.}
The same as Video Unshuffle.

%% file: appendices/task_details.tex
\clearpage
\section{Environment Episode Progression}
\label{appx:task_details}

Referenced in \cref{tab:task_overview} and \cref{sec:vlmgym}, this section presents detailed episode progressions for each environment. 
A summary index with page numbers is provided below:

\EnvEntry{Colorization}{subset:colorization}
\EnvEntry{Counting}{subsec:counting}
\EnvEntry{Jigsaw}{subsec:jigsaw}
\EnvEntry{Matchstick Equation}{subsec:matchstick_equation}
\EnvEntry{Matchstick Rotation}{subsec:matchstick_rotation}
\EnvEntry{Maze 2D}{subsec:maze2d}
\EnvEntry{Maze 3D}{subsec:maze3d}
\EnvEntry{Mental Rotation 2D}{subsec:mr2d}
\EnvEntry{Mental Rotation 3D \textsc{(Cube)}}{subsec:mr3d_cube}
\EnvEntry{Mental Rotation 3D \textsc{(Objaverse)}}{subsec:mr3d_objaverse}
\EnvEntry{MuJoCo Fetch \textsc{Pick-and-Place}}{subsec:mujoco_pick_place}
\EnvEntry{MuJoCo Fetch \textsc{Reach}}{subsec:mujoco_reach}
\EnvEntry{Patch Reassembly}{subsec:patch_reassembly}
\EnvEntry{Referring Dot-Pointing}{subsec:referring_dot}
\EnvEntry{Sliding Block}{subsec:sliding_block}
\EnvEntry{Video Unshuffle}{subsec:video_unshuffle}
\EnvEntry{Zoom-In Puzzle}{subsec:zoom_in_puzzle}

\vspace{0em}

%% file: appendices/ascii.tex
\section{ASCII-based Observation Visualization}
\label{appx:ascii_task_visualizations}

\begin{figure}[t]
\centering

\begin{minipage}[t]{0.48\linewidth}

    \begin{minipage}[t]{\linewidth}
        \centering
        \includegraphics[width=0.6\linewidth]{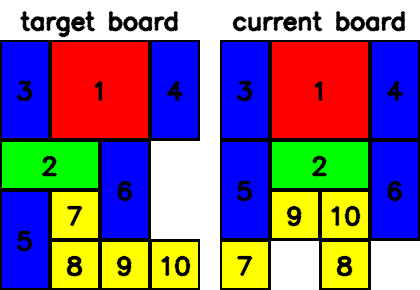}
        \caption*{Visual Rendering (default)}

\begin{lstlisting}[basicstyle=\ttfamily\small, frame=single]
Target Current
-----------
3114 | 3114
3114 | 3114
226. | 5226
576. | 5906
5890 | 7.8.  
\end{lstlisting}
        \vspace{-2ex}
        \caption*{Text Representation (variant)}
        \vspace{-2ex}
        \caption*{\textbf{Sliding Block}}
\noindent\rule{\linewidth}{0.4pt}

    \end{minipage}
    \vspace{2ex}

    \begin{minipage}[t]{\linewidth}
        \centering
        \includegraphics[width=0.6\linewidth]{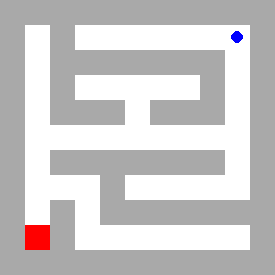}
        \caption*{Visual Rendering (default)}

\begin{lstlisting}[basicstyle=\ttfamily\scriptsize, frame=single]
#############
#############
## #      A##
## ####### ##
## #     # ##
## ### ### ##
##         ##
## ####### ##
##   #     ##
## # ########
##T#       ##
#############
#############
\end{lstlisting}
        \vspace{-2ex}
        \caption*{Text Representation (variant)}
        \vspace{-2ex}
        \caption*{\textbf{Maze 2D}}
    \end{minipage}

\end{minipage}
\hfill
\begin{minipage}[t]{0.48\linewidth}
    \begin{minipage}[t]{\linewidth}
        \centering
        \includegraphics[width=\linewidth]{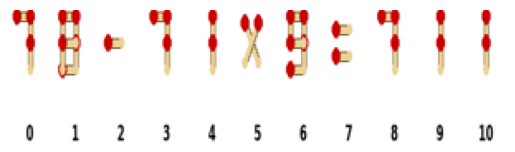}
        \vspace{-3ex}
        \caption*{Visual Rendering (default)}

\begin{lstlisting}[basicstyle=\fontsize{5.5}{6}\ttfamily, frame=single]
 ---   ---         ---               ---         ---              
    | |   |           |     |  \ /  |   |  ---      |     |     | 
    | |---|  ---      |     |   /   |---|           |     |     | 
    | |   |           |     |  / \      |           |     |     | 
    | |   |           |     |           |  ---      |     |     | 
       ---                           ---                          
  0     1     2     3     4     5     6     7     8     9    10   
\end{lstlisting}
        \vspace{-2ex}
        \caption*{Text Representation (variant)}
        \vspace{-2ex}
        \caption*{\textbf{Matchstick Equation}}
        \vspace{1ex}
    \end{minipage}

\noindent\rule{\linewidth}{0.4pt}

    \begin{minipage}[t]{\linewidth}
        \centering
        \includegraphics[width=0.5\linewidth]{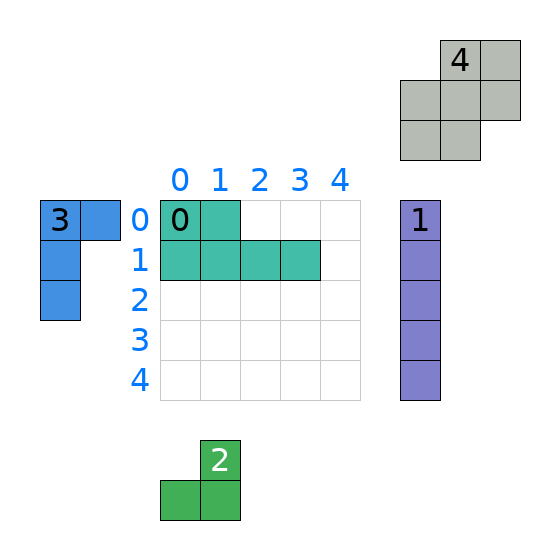}
        \vspace{-2ex}
        \caption*{Visual Rendering (default)}

\begin{lstlisting}[basicstyle=\ttfamily\scriptsize, frame=single]
   0  1  2  3  4 
  ---------------
0 | 0  0  .  .  . 
1 | 0  0  0  0  . 
2 | .  .  .  .  . 
3 | .  .  .  .  . 
4 | .  .  .  .  . 

--- Parked Patches ---

Patch 1:
  *
  1
  1
  1
  1

Patch 2:
   *
  22

Patch 3:
  *3
  3 
  3 

Patch 4:
   *4
  444
  44
\end{lstlisting}
        \vspace{-2ex}
        \caption*{Text Representation (variant)}
        \vspace{-2ex}
        \caption*{\textbf{Patch Reassembly}}
    \end{minipage}

\end{minipage}
\vspace{-2ex}
\caption{Visual and text representations across four environments}
\label{fig:text_mode_all}
\end{figure}

In this section, we present example episode variants rendered in text, as discussed in \cref{subsec:text_repr}, for the Sliding Block, Maze 2D, Patch Reassembly, and Matchstick Equation environments in \cref{fig:text_mode_all}.

Note that the instructions are slightly adapted to fit the text-based format (\textit{e.g.}, in the visual version of Patch Reassembly, we describe the \textit{anchor} as “the cell that shows the patch’s ID number,” while in the text version we note that “the anchor cell for each parked patch is marked with a `*' instead of its ID number”).

\section{\ours Interface}
\label{appx:interface}
\input{algs/step}

In this section, we present the pseudocode for the \texttt{step} function (\cref{alg:step}) used in \ours (\textit{i.e.,}  \cref{sec:vlmgym}). The function initializes the reward and both termination flags, then parses the model’s output string into an action name and payload. If parsing fails, it immediately returns an observation with ``invalid format’’ as feedback.

If the parsed action name is supported and its payload is valid for the corresponding action space, the function calls \texttt{Apply}, which executes the action and returns the environment feedback. Otherwise, it ends early with ``invalid action’’ as feedback.

Termination and truncation are determined inside \texttt{Apply}. If the action triggers termination (\textit{e.g.}, \texttt{stop}), the function computes the final reward based on the environment state. Thus, the returned reward is always zero for non-terminal transitions and the final score upon termination.

Finally, the function returns the new observation, reward, termination, and truncation flags, and the feedback describing the action outcome.

%% file: algs/step.tex
\begin{wrapfigure}{r}{0.35\linewidth}
\vspace{-5ex}
\begin{minipage}{\linewidth}
\begin{algorithm}[H]
\small
\caption{\small{Generic Step Function (\cref{appx:interface}). 
Symbols: $\rho$ = reward, $\tau$ = terminated, $\upsilon$ = truncated,
$\varphi$ = feedback, $\alpha$ = action name, $\pi$ = payload, $\iota$ = info.}}
\label{alg:step}
\begin{algorithmic}
\Function{Step}{$a$}
  \State $\rho \gets 0$
  \State $(\tau,\upsilon) \gets (false,false)$
  \State Parse $a \to (\alpha,\pi)$

  \If{invalid format}
    \State \Return $(obs(),0,\tau,\upsilon,\iota(\text{``invalid format''}))$
  \EndIf

  \If{$\alpha \in \mathcal{A}$ and $\pi \in \mathcal{A}[\alpha]$}
    \State $(\varphi,\tau,\upsilon) \gets$ Apply$(\alpha,\pi)$
  \Else
    \State \Return $(obs(),0,\tau,\upsilon,\iota(\text{``invalid action''}))$
  \EndIf

  \If{$\tau = true$}
    \State $\rho \gets$ ComputeReward()
  \EndIf

  \State \Return $(obs(),\rho,\tau,\upsilon,\iota(\varphi))$
\EndFunction
\end{algorithmic}
\end{algorithm}
\end{minipage}
\vspace{-2ex}
\end{wrapfigure}

%% file: appendices/interface_better.tex
\section{Configuration of Environments}
\label{appx:task_variant_generalization}
In this section, we provide the tunable parameters for each task that determine its difficulty.
Note that in some environments, higher difficulty primarily increases the complexity of
the input observations (e.g., \cref{subsec:base_generalization}), while in others it tightens the reward criteria and requires more
precise control. Details are in \cref{tab:task_config_full}.

\vspace{3ex}

\begin{table*}[h] %
\centering
\small
\captionof{table}{\textbf{Configuration summary for all tasks}: tunable parameters, easy and hard configurations, and source datasets.}
\label{tab:task_config_full}

\centering
\small
\begin{tabular}{@{}p{0.15\linewidth} p{0.36\linewidth} p{0.13\linewidth} p{0.13\linewidth} p{0.12\linewidth}@{}}
\toprule
\textbf{Task} &
\textbf{Tunable Difficulty Parameters} &
\textbf{Easy} &
\textbf{Hard} &
\textbf{Src. Dataset} \\
\midrule

Colorization &
Accuracy radius \texttt{ar} (precision required for hue and saturation match). &
\texttt{ar} = 11 &
\texttt{ar} = 16 &
LLaVA \cite{liu2023llava} \\

\midrule
Counting &
Minimum and maximum \newline count range \texttt{c\_min}, \texttt{c\_max}. &
\texttt{c\_min} = 2, \newline \texttt{c\_max} = 20 &
\texttt{c\_min} = 5, \newline \texttt{c\_max} = 30 &
LVIS \cite{gupta2019lvis} \\

\midrule
Jigsaw &
number of rows and columns \texttt{nr}, \texttt{nc}. &
\texttt{nr} = 2, \texttt{nc} = 2 &
\texttt{nr} = 3, \texttt{nc} = 3 &
LLaVA \cite{liu2023llava} \\

\midrule
Matchstick \newline Equation &
Number of break moves\newline \texttt{bm} (corruptions to fix). &
\texttt{bm} = 1 &
\texttt{bm} = 2 &
--- \\

\midrule
Matchstick\newline Rotation &
Hidden scale range \texttt{sr}, position tolerance \texttt{pt}, angular tolerance \texttt{at}. &
\texttt{pt} = 10, \newline\texttt{at} = 15 &
\texttt{pt} = 5, \newline\texttt{at} = 10 &
--- \\

\midrule
Maze 2D &
Maze width and height \texttt{mw}, \texttt{mh}. &
\texttt{mw} = 9, \texttt{mh} = 9 &
\texttt{mw} = 11, \texttt{mh} = 11 &
--- \\

\midrule
Maze 3D &
Maze width and height \texttt{mw}, \texttt{mh}. &
\texttt{mw} = 7, \texttt{mh} = 7 &
\texttt{mw} = 9, \texttt{mh} = 9 &
--- \\

\midrule
Mental Rotation 2D &
Angular tolerance \texttt{at}. &
\texttt{at} = 10.0 &
\texttt{at} = 5.0 &
LLaVA \cite{liu2023llava} \\

\midrule
Mental Rotation 3D \newline \textsc{(Cube)} &
Number of segments \texttt{ns}, length range \texttt{lr}, angular tolerance \texttt{at}. &
\texttt{ns} = 4 &
\texttt{ns} = 6 &
--- \\

\midrule
Mental Rotation 3D \newline \textsc{(Objaverse)} &
Angular tolerance \texttt{at}. &
\texttt{at} = 15.0 &
\texttt{at} = 5.0 &
Objaverse \cite{objaverse} \\

\midrule
MuJoCo Fetch \newline \textsc{Pick-and-Place} &
No user-tuned difficulty parameters \newline(standardized task). &
Standard &
Standard &
--- \\

\midrule
MuJoCo Fetch \newline \textsc{Reach} &
No user-tuned difficulty parameters \newline(standardized task). &
Standard &
Standard &
--- \\

\midrule
Patch Reassembly &
Grid size \texttt{gs}, number of patches \texttt{np}. &
\texttt{gs} = (6, 6),\newline \texttt{np} = 5 &
\texttt{gs} = (8, 8),\newline \texttt{np} = 6 &
--- \\

\midrule
Referring \newline \textsc{Dot-Pointing} &
No user-tuned difficulty parameters \newline(standardized task). &
Standard &
Standard &
RefCOCO \cite{kazemzadeh-etal-2014-referitgame} \\

\midrule
Sliding Block &
Number of shuffle moves \texttt{sm}. &
\texttt{sm} = 30 &
\texttt{sm} = 90 &
--- \\

\midrule
Video Unshuffle &
Number of frames \texttt{nf}, sampling strategy \texttt{ss}, minimum frame-diff threshold \texttt{mfd}. &
\texttt{nf} = 4 &
\texttt{nf} = 5 &
SS2 \cite{goyal2017something} \\

\midrule
Zoom-In Puzzle &
Zoom gap \texttt{zg}, zoom variability \texttt{zs}, minimum zoom \texttt{mz}, num. of views \texttt{zv}, nested crop \texttt{nest}. &
\texttt{zv} = 4 &
\texttt{zv} = 5 &
LLaVA \cite{liu2023llava} \\

\bottomrule
\end{tabular}
\end{table*}

%% file: tables/stringsight.tex
\begin{table*}[t]
\centering
\small
\captionof{table}{\textbf{Example clusters discovered by StringSight.}}
\label{tab:stringsight_discovery_clusters}
\vspace{-0.5em}
\centering
\small
\begin{tabular}{@{} p{0.03\linewidth} p{0.93\linewidth} @{}}
\toprule
 & \textbf{Cluster Description} \\
\midrule

\multicolumn{2}{l}{\textbf{MuJoCo Fetch (Pick-and-Place)}} \\

\textbullet & The model issues repetitive movement commands without adapting based on task progress or environment feedback, resulting in oscillatory behaviors like moving up and down, left and right, or in a single direction with no meaningful progress toward grasping or placing the cube. For example, it may move forward repeatedly even after overshooting the target or oscillate without ever attempting a grasp. \\

\textbullet & Ignores visual feedback from images and does not adjust its actions in response to clear cues about the state or position of the cube, gripper, or target marker. This results in the model issuing irrelevant or counterproductive actions, such as continuing to move when the cube has not been grasped. \\

\textbullet & Issues the “stop” command prematurely before the end-effector is even close to the target marker, ending the task early without justification or clear success criteria. \\
\midrule

\multicolumn{2}{l}{\textbf{Mental Rotation 3D (Cube)}} \\

\textbullet & The model repeatedly issues the same rotation action, often on a single axis, for many steps without adapting based on visual feedback, resulting in rigid loops or unbroken patterns. \\

\textbullet & The model oscillates between a small set of orientations, alternating or repeating similar rotations (e.g., $+45^{\circ}$, $-45^{\circ}$), causing the object to cycle endlessly without making progress toward the target alignment. \\
\midrule

\multicolumn{2}{l}{\textbf{Zoom-In Puzzle}} \\

\textbullet & The model finalizes the arrangement immediately without performing any swaps, reordering, or verification, accepting the initial sequence as correct regardless of its accuracy. For example, it issues a “stop” command on the first step even if the order is incorrect. \\
\midrule

\multicolumn{2}{l}{\textbf{Matchstick Rotation}} \\

\textbullet & The model issues fixed or monotonically decreasing movement and rotation magnitudes without attempting to estimate the unknown scale or using exploratory actions to resolve scale ambiguity, proceeding as if the appropriate step size is already known. \\

\textbullet & Action sequences do not adapt based on feedback or observed outcomes; the model follows a predetermined or repetitive strategy without checking if moves are effective or responding to evidence from the environment. \\

\midrule

\multicolumn{2}{l}{\textbf{Maze 2D}} \\

\textbullet & The model repeatedly issues the same invalid movement commands, such as trying to move into walls, even after receiving explicit feedback that these actions are not possible. For example, it continues to try moving left into a wall after being told each time that the move is blocked. \\

\textbullet & Does not build, update, or use any internal map or memory of previous moves or environmental feedback, resulting in repeated visits to the same locations, blocked paths, and inefficient looping navigation. \\
\midrule

\multicolumn{2}{l}{\textbf{Maze 3D}} \\

\textbullet & Movement decisions are based exclusively on immediate sensory feedback, without building or referencing internal memory or a map of previously explored locations, leading to repeated wall collisions, revisiting dead ends, and inefficient navigation. \\

\textbullet & Frequently issues long sequences of consecutive turning actions—such as alternating left and right—without forward movement or meaningful progress toward the goal, resulting in wasted steps and inefficient pathfinding. \\

\bottomrule
\end{tabular}
\vspace{-1em}
\end{table*}

%% file: figs/tast_difficulty_ranking.tex
\begin{figure*}[h]
\includegraphics[width=\linewidth]{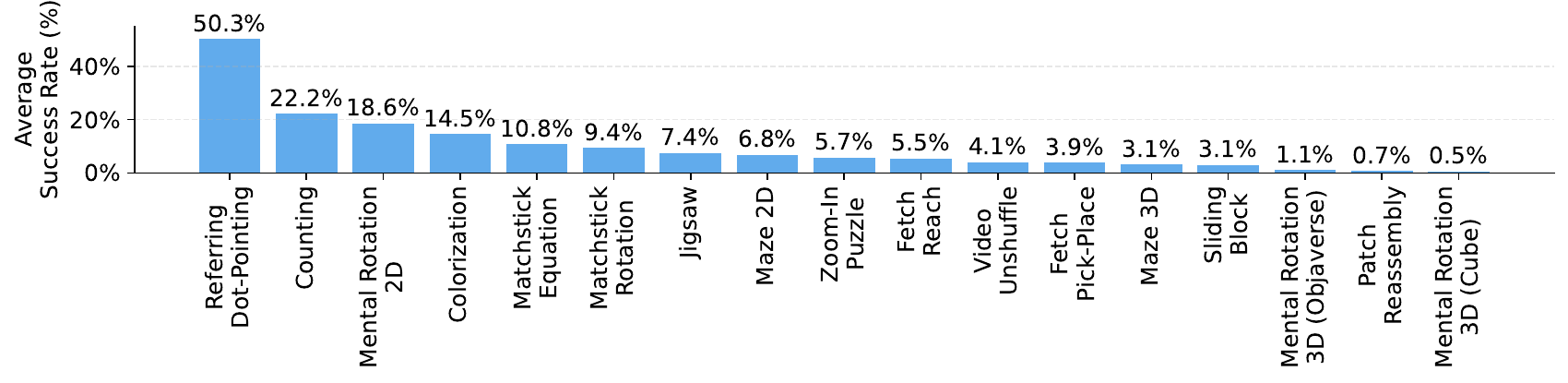}
\vspace{-5ex}
    \caption{\textbf{Average success rate across frontier models on each task}. The easiest tasks are Referring Dot-Pointing, and Counting, with over 20\% accuracy on average across all models, while the hardest tasks are Mental Rotation 3D (Cube), Patch Reassembly, and Mental Rotation, with the average accuracy less than 2\% on average. \colin{Averaged -\> Average Across}}
    \label{fig:task_difficulty_ranking}
    \vspace{-2.5ex}
\end{figure*}

%% file: figs/num_steps.tex
\begin{figure*}[t]
\includegraphics[width=\linewidth]{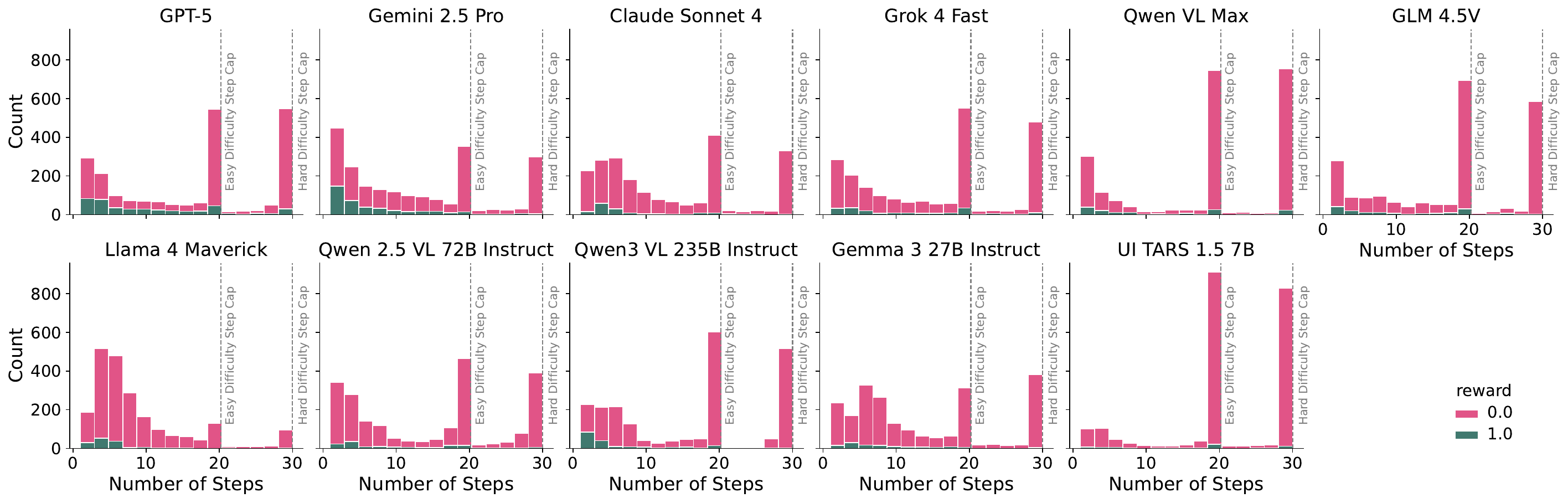}
\vspace{-4ex}
    \caption{\textbf{The Number of Steps each Model Takes Over all Tasks}. We calculate the number of steps over all trajectories for each model and visualize the correct trajectories (green) and the incorrect trajectories (red).}
    \label{fig:num_steps}
\end{figure*}

%% file: figs/easy_vs_hard.tex
\begin{wrapfigure}{r}{0.6\linewidth}
    \centering
    \includegraphics[width=\linewidth, trim={0mm 0mm 0mm 0mm}, clip]{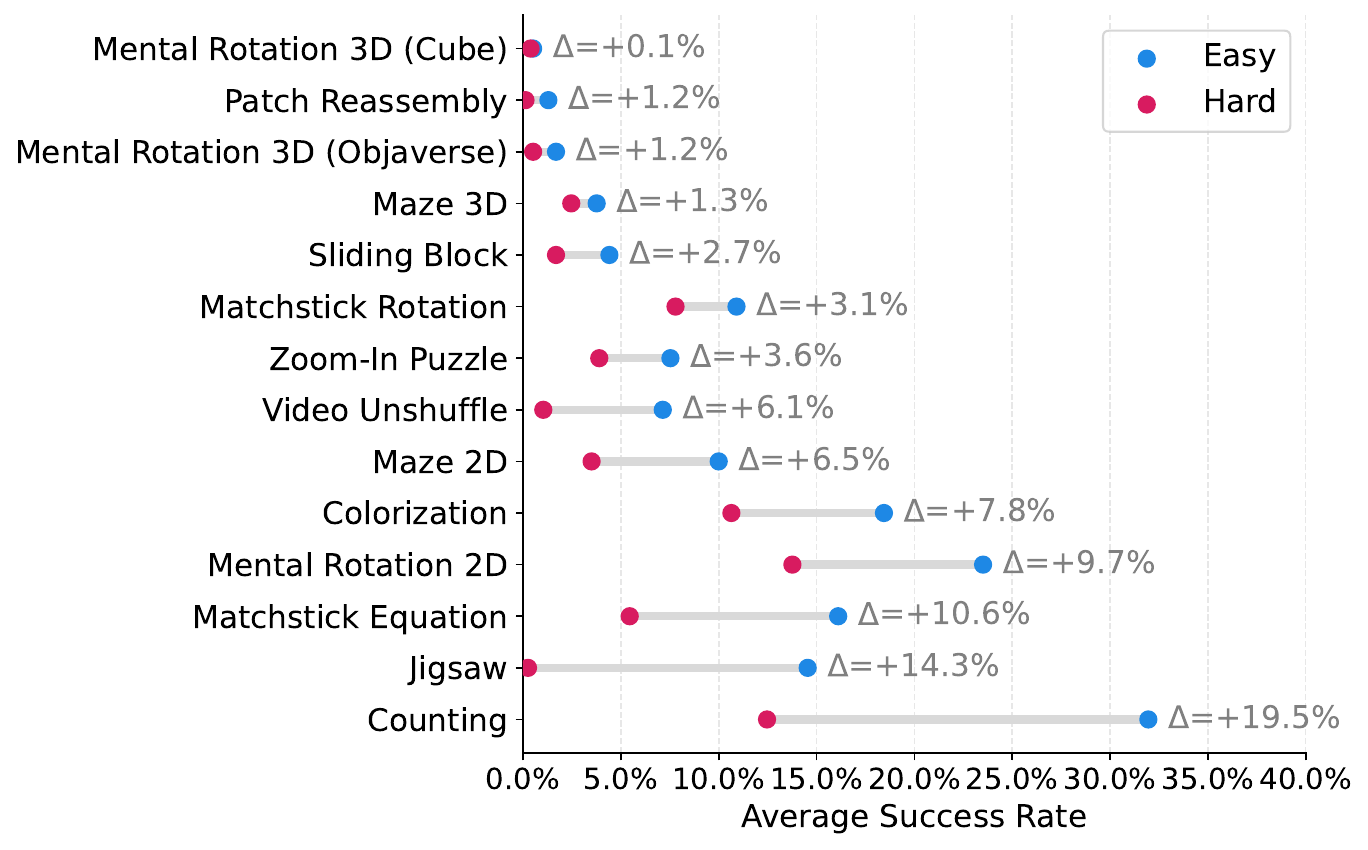}
    \vspace{-3ex}
    \caption{\textbf{Easy $\rightarrow$ Hard Performance Drop}. For each task, we calculate the average accuracy on Easy and Hard, respectively, over all models, and then visualize the performance drop between Easy and Hard.}
    \label{fig:easy_hard}
    \vspace{-3ex}
\end{wrapfigure}

%% file: main.bbl
\begin{thebibliography}{108}
\providecommand{\natexlab}[1]{#1}
\providecommand{\url}[1]{\texttt{#1}}
\expandafter\ifx\csname urlstyle\endcsname\relax
  \providecommand{\doi}[1]{doi: #1}\else
  \providecommand{\doi}{doi: \begingroup \urlstyle{rm}\Url}\fi

\bibitem[Abdulhai et~al.(2023)Abdulhai, White, Snell, Sun, Hong, Zhai, Xu, and Levine]{abdulhai2023lmrl}
Marwa Abdulhai, Isadora White, Charlie Snell, Charles Sun, Joey Hong, Yuexiang Zhai, Kelvin Xu, and Sergey Levine.
\newblock Lmrl gym: Benchmarks for multi-turn reinforcement learning with language models.
\newblock \emph{arXiv preprint arXiv:2311.18232}, 2023.

\bibitem[Achiam et~al.(2023)Achiam, Adler, Agarwal, Ahmad, Akkaya, Aleman, Almeida, Altenschmidt, Altman, Anadkat, et~al.]{achiam2023gpt}
Josh Achiam, Steven Adler, Sandhini Agarwal, Lama Ahmad, Ilge Akkaya, Florencia~Leoni Aleman, Diogo Almeida, Janko Altenschmidt, Sam Altman, Shyamal Anadkat, et~al.
\newblock Gpt-4 technical report.
\newblock \emph{arXiv preprint arXiv:2303.08774}, 2023.

\bibitem[Ahmed et~al.(2020)Ahmed, Tr{\"a}uble, Goyal, Neitz, Bengio, Sch{\"o}lkopf, W{\"u}thrich, and Bauer]{ahmed2020causalworld}
Ossama Ahmed, Frederik Tr{\"a}uble, Anirudh Goyal, Alexander Neitz, Yoshua Bengio, Bernhard Sch{\"o}lkopf, Manuel W{\"u}thrich, and Stefan Bauer.
\newblock Causalworld: A robotic manipulation benchmark for causal structure and transfer learning.
\newblock \emph{arXiv preprint arXiv:2010.04296}, 2020.

\bibitem[Alayrac et~al.(2022)Alayrac, Donahue, Luc, Miech, Barr, Hasson, Lenc, Mensch, Millican, Reynolds, et~al.]{alayrac2022flamingo}
Jean-Baptiste Alayrac, Jeff Donahue, Pauline Luc, Antoine Miech, Iain Barr, Yana Hasson, Karel Lenc, Arthur Mensch, Katherine Millican, Malcolm Reynolds, et~al.
\newblock Flamingo: a visual language model for few-shot learning.
\newblock \emph{Advances in neural information processing systems}, 35:\penalty0 23716--23736, 2022.

\bibitem[Bai et~al.(2025)Bai, Chen, Liu, Wang, Ge, Song, Dang, Wang, Wang, Tang, et~al.]{bai2025qwen2}
Shuai Bai, Keqin Chen, Xuejing Liu, Jialin Wang, Wenbin Ge, Sibo Song, Kai Dang, Peng Wang, Shijie Wang, Jun Tang, et~al.
\newblock Qwen2. 5-vl technical report.
\newblock \emph{arXiv preprint arXiv:2502.13923}, 2025.

\bibitem[Baird(1970)]{Baird1970}
John~C. Baird.
\newblock \emph{Psychophysical Analysis of Visual Space}.
\newblock International Series of Monographs in Experimental Psychology. Pergamon Press / Elsevier, 1970.
\newblock ISBN 978-0-08-013876-3.
\newblock URL \url{https://www.sciencedirect.com/book/9780080138763/psychophysical-analysis-of-visual-space}.

\bibitem[Beattie et~al.(2016)Beattie, Leibo, Teplyashin, Ward, Wainwright, K{\"u}ttler, Lefrancq, Green, Vald{\'e}s, Sadik, et~al.]{beattie2016deepmind}
Charles Beattie, Joel~Z Leibo, Denis Teplyashin, Tom Ward, Marcus Wainwright, Heinrich K{\"u}ttler, Andrew Lefrancq, Simon Green, V{\'\i}ctor Vald{\'e}s, Amir Sadik, et~al.
\newblock Deepmind lab.
\newblock \emph{arXiv preprint arXiv:1612.03801}, 2016.

\bibitem[{Bellemare} et~al.(2013){Bellemare}, {Naddaf}, {Veness}, and {Bowling}]{bellemare13arcade}
M.~G. {Bellemare}, Y.~{Naddaf}, J.~{Veness}, and M.~{Bowling}.
\newblock The arcade learning environment: An evaluation platform for general agents.
\newblock \emph{Journal of Artificial Intelligence Research}, 47:\penalty0 253--279, jun 2013.

\bibitem[Bjorck et~al.(2025)Bjorck, Casta{\~n}eda, Cherniadev, Da, Ding, Fan, Fang, Fox, Hu, Huang, et~al.]{bjorck2025gr00t}
Johan Bjorck, Fernando Casta{\~n}eda, Nikita Cherniadev, Xingye Da, Runyu Ding, Linxi Fan, Yu~Fang, Dieter Fox, Fengyuan Hu, Spencer Huang, et~al.
\newblock Gr00t n1: An open foundation model for generalist humanoid robots.
\newblock \emph{arXiv preprint arXiv:2503.14734}, 2025.

\bibitem[Black et~al.(2024)Black, Brown, Driess, Esmail, Equi, Finn, Fusai, Groom, Hausman, Ichter, et~al.]{black2410pi0}
Kevin Black, Noah Brown, Danny Driess, Adnan Esmail, Michael Equi, Chelsea Finn, Niccolo Fusai, Lachy Groom, Karol Hausman, Brian Ichter, et~al.
\newblock $\pi$0: A vision-language-action flow model for general robot control. corr, abs/2410.24164, 2024. doi: 10.48550.
\newblock \emph{arXiv preprint ARXIV.2410.24164}, 2024.

\bibitem[Brockman et~al.(2016)Brockman, Cheung, Pettersson, Schneider, Schulman, Tang, and Zaremba]{2016gym}
Greg Brockman, Vicki Cheung, Ludwig Pettersson, Jonas Schneider, John Schulman, Jie Tang, and Wojciech Zaremba.
\newblock Openai gym, 2016.

\bibitem[Caccia et~al.(2024)Caccia, Thakkar, Boisvert, de~Chezelles, Pich{\'e}, Chapados, Drouin, Gasse, and Lacoste]{caccia2024finetuning}
Massimo Caccia, Megh Thakkar, L{\'e}o Boisvert, Thibault Le~Sellier de~Chezelles, Alexandre Pich{\'e}, Nicolas Chapados, Alexandre Drouin, Maxime Gasse, and Alexandre Lacoste.
\newblock Fine-tuning web agents: It works, but it's trickier than you think.
\newblock In \emph{NeurIPS 2024 Workshop on Open-World Agents}, 2024.
\newblock URL \url{https://openreview.net/forum?id=SkwtxEkst2}.

\bibitem[Cao et~al.(2025)Cao, Mckenna, Graf, and Oyekan]{cao2025learn}
Guanqun Cao, Ryan Mckenna, Erich Graf, and John Oyekan.
\newblock Learn from the past: Language-conditioned object rearrangement with large language models.
\newblock \emph{arXiv preprint arXiv:2501.18516}, 2025.

\bibitem[Chen et~al.(2025)Chen, Gao, Liu, Huang, Sung, Zhou, Wu, and Chang]{chen2025g1}
Liang Chen, Hongcheng Gao, Tianyu Liu, Zhiqi Huang, Flood Sung, Xinyu Zhou, Yuxin Wu, and Baobao Chang.
\newblock G1: Bootstrapping perception and reasoning abilities of vision-language model via reinforcement learning.
\newblock \emph{arXiv preprint arXiv:2505.13426}, 2025.

\bibitem[Chen et~al.(2024)Chen, Li, Dong, Zhang, Zang, Chen, Duan, Wang, Qiao, Lin, et~al.]{chen2024we}
Lin Chen, Jinsong Li, Xiaoyi Dong, Pan Zhang, Yuhang Zang, Zehui Chen, Haodong Duan, Jiaqi Wang, Yu~Qiao, Dahua Lin, et~al.
\newblock Are we on the right way for evaluating large vision-language models?
\newblock \emph{Advances in Neural Information Processing Systems}, 37:\penalty0 27056--27087, 2024.

\bibitem[Choi et~al.(2024)Choi, Yoon, Ong, Kim, and Jang]{choi2024lota}
Jae-Woo Choi, Youngwoo Yoon, Hyobin Ong, Jaehong Kim, and Minsu Jang.
\newblock Lota-bench: Benchmarking language-oriented task planners for embodied agents.
\newblock \emph{arXiv preprint arXiv:2402.08178}, 2024.

\bibitem[Coelho et~al.(2025)Coelho, Ning, He, Mao, Paladugu, Setlur, Jin, Callan, Magalh{\~a}es, Martins, et~al.]{coelho2025deepresearchgym}
Jo{\~a}o Coelho, Jingjie Ning, Jingyuan He, Kangrui Mao, Abhijay Paladugu, Pranav Setlur, Jiahe Jin, Jamie Callan, Jo{\~a}o Magalh{\~a}es, Bruno Martins, et~al.
\newblock Deepresearchgym: A free, transparent, and reproducible evaluation sandbox for deep research.
\newblock \emph{arXiv preprint arXiv:2505.19253}, 2025.

\bibitem[Cooper and Shepard(1973)]{COOPER197375}
Lynn~A. Cooper and Roger~N. Shepard.
\newblock Chronometric studies of the rotation of mental images.
\newblock In WILLIAM~G. CHASE, editor, \emph{Visual Information Processing}, pages 75--176. Academic Press, 1973.
\newblock ISBN 978-0-12-170150-5.
\newblock \doi{https://doi.org/10.1016/B978-0-12-170150-5.50009-3}.
\newblock URL \url{https://www.sciencedirect.com/science/article/pii/B9780121701505500093}.

\bibitem[DeepMind(2025)]{GoogleDeepMind2025_Gemini2_5Pro}
Google DeepMind.
\newblock {Gemini 2.5 Pro Technical Report}.
\newblock Technical report, Google DeepMind, 2025.
\newblock URL \url{https://modelcards.withgoogle.com/assets/documents/gemini-2.5-pro.pdf}.
\newblock Accessed: 2025-11-06.

\bibitem[Deitke et~al.(2023)Deitke, Schwenk, Salvador, Weihs, Michel, VanderBilt, Schmidt, Ehsani, Kembhavi, and Farhadi]{objaverse}
Matt Deitke, Dustin Schwenk, Jordi Salvador, Luca Weihs, Oscar Michel, Eli VanderBilt, Ludwig Schmidt, Kiana Ehsani, Aniruddha Kembhavi, and Ali Farhadi.
\newblock Objaverse: A universe of annotated 3d objects.
\newblock In \emph{Proceedings of the IEEE/CVF conference on computer vision and pattern recognition}, pages 13142--13153, 2023.

\bibitem[Deng et~al.(2023)Deng, Gu, Zheng, Chen, Stevens, Wang, Sun, and Su]{deng2023mind2web}
Xiang Deng, Yu~Gu, Boyuan Zheng, Shijie Chen, Sam Stevens, Boshi Wang, Huan Sun, and Yu~Su.
\newblock Mind2web: Towards a generalist agent for the web.
\newblock \emph{Advances in Neural Information Processing Systems}, 36:\penalty0 28091--28114, 2023.

\bibitem[Dubey et~al.(2024)Dubey, Jauhri, Pandey, Kadian, Al-Dahle, Letman, Mathur, Schelten, Yang, Fan, et~al.]{dubey2024llama}
Abhimanyu Dubey, Abhinav Jauhri, Abhinav Pandey, Abhishek Kadian, Ahmad Al-Dahle, Aiesha Letman, Akhil Mathur, Alan Schelten, Amy Yang, Angela Fan, et~al.
\newblock The llama 3 herd of models.
\newblock \emph{arXiv e-prints}, pages arXiv--2407, 2024.

\bibitem[Dunlap et~al.(2025)Dunlap, Mandal, Darrell, Steinhardt, and Gonzalez]{dunlap2025vibecheck}
Lisa Dunlap, Krishna Mandal, Trevor Darrell, Jacob Steinhardt, and Joseph~E Gonzalez.
\newblock Vibecheck: Discover and quantify qualitative differences in large language models.
\newblock In \emph{International Conference on Learning Representations (ICLR)}, 2025.
\newblock URL \url{https://lisadunlap.github.io/VibeCheck/}.

\bibitem[Ehsani et~al.(2021)Ehsani, Han, Herrasti, VanderBilt, Weihs, Kolve, Kembhavi, and Mottaghi]{ehsani2021manipulathor}
Kiana Ehsani, Winson Han, Alvaro Herrasti, Eli VanderBilt, Luca Weihs, Eric Kolve, Aniruddha Kembhavi, and Roozbeh Mottaghi.
\newblock Manipulathor: A framework for visual object manipulation.
\newblock In \emph{Proceedings of the IEEE/CVF conference on computer vision and pattern recognition}, pages 4497--4506, 2021.

\bibitem[Fujii et~al.(1998)Fujii, Inui, Tokunaga, and Tanaka]{fujii-etal-1998-selective}
Atsushi Fujii, Kentaro Inui, Takenobu Tokunaga, and Hozumi Tanaka.
\newblock Selective sampling for example-based word sense disambiguation.
\newblock \emph{Computational Linguistics}, 24\penalty0 (4):\penalty0 573--597, 1998.
\newblock URL \url{https://aclanthology.org/J98-4002/}.

\bibitem[Gibson(1979)]{Gibson1979TheEA}
James~J. Gibson.
\newblock \emph{The Ecological Approach to Visual Perception}.
\newblock Houghton Mifflin, Boston, MA, 1979.
\newblock ISBN 0-395-27049-9.

\bibitem[Gidaris et~al.(2018)Gidaris, Singh, and Komodakis]{gidaris2018unsupervised}
Spyros Gidaris, Praveer Singh, and Nikos Komodakis.
\newblock Unsupervised representation learning by predicting image rotations.
\newblock \emph{arXiv preprint arXiv:1803.07728}, 2018.

\bibitem[Golomb(1994)]{Golomb1994Polyominoes}
Solomon~W. Golomb.
\newblock \emph{Polyominoes: Puzzles, Patterns, Problems, and Packings}.
\newblock Princeton University Press, Princeton, NJ, 2nd edition, 1994.
\newblock ISBN 978-0-691-02444-8.

\bibitem[Goyal et~al.(2017)Goyal, Ebrahimi~Kahou, Michalski, Materzynska, Westphal, Kim, Haenel, Fruend, Yianilos, Mueller-Freitag, et~al.]{goyal2017something}
Raghav Goyal, Samira Ebrahimi~Kahou, Vincent Michalski, Joanna Materzynska, Susanne Westphal, Heuna Kim, Valentin Haenel, Ingo Fruend, Peter Yianilos, Moritz Mueller-Freitag, et~al.
\newblock The" something something" video database for learning and evaluating visual common sense.
\newblock In \emph{Proceedings of the IEEE international conference on computer vision}, pages 5842--5850, 2017.

\bibitem[Gupta et~al.(2019)Gupta, Dollar, and Girshick]{gupta2019lvis}
Agrim Gupta, Piotr Dollar, and Ross Girshick.
\newblock Lvis: A dataset for large vocabulary instance segmentation.
\newblock In \emph{Proceedings of the IEEE/CVF conference on computer vision and pattern recognition}, pages 5356--5364, 2019.

\bibitem[Henderson(2001)]{Henderson2001ASA}
John~M. Henderson.
\newblock A sensorimotor account of vision and visual consciousness.
\newblock \emph{The Behavioral and brain sciences}, 24 5:\penalty0 939--73; discussion 973--1031, 2001.
\newblock URL \url{https://api.semanticscholar.org/CorpusID:22606536}.

\bibitem[Hong et~al.(2025)Hong, Yu, Gu, Wang, Gan, Tang, Cheng, Qi, Ji, Pan, et~al.]{hong2025glm}
Wenyi Hong, Wenmeng Yu, Xiaotao Gu, Guo Wang, Guobing Gan, Haomiao Tang, Jiale Cheng, Ji~Qi, Junhui Ji, Lihang Pan, et~al.
\newblock Glm-4.1 v-thinking: Towards versatile multimodal reasoning with scalable reinforcement learning.
\newblock \emph{arXiv e-prints}, pages arXiv--2507, 2025.

\bibitem[Hu et~al.(2025)Hu, Huo, Zhang, Yu, Xing, Stoica, Rosing, Jin, and Zhang]{hu2025lmgame}
Lanxiang Hu, Mingjia Huo, Yuxuan Zhang, Haoyang Yu, Eric~P Xing, Ion Stoica, Tajana Rosing, Haojian Jin, and Hao Zhang.
\newblock lmgame-bench: How good are llms at playing games?
\newblock \emph{arXiv preprint arXiv:2505.15146}, 2025.

\bibitem[Huang et~al.(2025)Huang, Liu, Fu, Wu, Mukadam, Malik, Goldberg, and Abbeel]{huang2025otter}
Huang Huang, Fangchen Liu, Letian Fu, Tingfan Wu, Mustafa Mukadam, Jitendra Malik, Ken Goldberg, and Pieter Abbeel.
\newblock Otter: A vision-language-action model with text-aware feature extraciton.
\newblock \emph{arXiv preprint arXiv:2503.03734}, 2025.

\bibitem[James et~al.(2020)James, Ma, Arrojo, and Davison]{james2020rlbench}
Stephen James, Zicong Ma, David~Rovick Arrojo, and Andrew~J Davison.
\newblock Rlbench: The robot learning benchmark \& learning environment.
\newblock \emph{IEEE Robotics and Automation Letters}, 5\penalty0 (2):\penalty0 3019--3026, 2020.

\bibitem[Jang et~al.(2022)Jang, Irpan, Khansari, Kappler, Ebert, Lynch, Levine, and Finn]{jang2022bc}
Eric Jang, Alex Irpan, Mohi Khansari, Daniel Kappler, Frederik Ebert, Corey Lynch, Sergey Levine, and Chelsea Finn.
\newblock Bc-z: Zero-shot task generalization with robotic imitation learning.
\newblock In \emph{Conference on Robot Learning}, pages 991--1002. PMLR, 2022.

\bibitem[Jang et~al.(2024)Jang, Li, Zhao, Ding, Lin, Liang, Bonatti, and Koishida]{jang2024videowebarena}
Lawrence Jang, Yinheng Li, Dan Zhao, Charles Ding, Justin Lin, Paul~Pu Liang, Rogerio Bonatti, and Kazuhito Koishida.
\newblock Videowebarena: Evaluating long context multimodal agents with video understanding web tasks.
\newblock \emph{arXiv preprint arXiv:2410.19100}, 2024.

\bibitem[Jimenez et~al.(2023)Jimenez, Yang, Wettig, Yao, Pei, Press, and Narasimhan]{jimenez2023swe}
Carlos~E Jimenez, John Yang, Alexander Wettig, Shunyu Yao, Kexin Pei, Ofir Press, and Karthik Narasimhan.
\newblock Swe-bench: Can language models resolve real-world github issues?
\newblock \emph{arXiv preprint arXiv:2310.06770}, 2023.

\bibitem[Kazemzadeh et~al.(2014)Kazemzadeh, Ordonez, Matten, and Berg]{kazemzadeh-etal-2014-referitgame}
Sahar Kazemzadeh, Vicente Ordonez, Mark Matten, and Tamara Berg.
\newblock {R}efer{I}t{G}ame: Referring to objects in photographs of natural scenes.
\newblock In Alessandro Moschitti, Bo~Pang, and Walter Daelemans, editors, \emph{Proceedings of the 2014 Conference on Empirical Methods in Natural Language Processing ({EMNLP})}, pages 787--798, Doha, Qatar, October 2014. Association for Computational Linguistics.
\newblock \doi{10.3115/v1/D14-1086}.
\newblock URL \url{https://aclanthology.org/D14-1086/}.

\bibitem[Khanna et~al.(2024)Khanna, Ramrakhya, Chhablani, Yenamandra, Gervet, Chang, Kira, Chaplot, Batra, and Mottaghi]{khanna2024goat}
Mukul Khanna, Ram Ramrakhya, Gunjan Chhablani, Sriram Yenamandra, Theophile Gervet, Matthew Chang, Zsolt Kira, Devendra~Singh Chaplot, Dhruv Batra, and Roozbeh Mottaghi.
\newblock Goat-bench: A benchmark for multi-modal lifelong navigation.
\newblock In \emph{Proceedings of the IEEE/CVF Conference on Computer Vision and Pattern Recognition}, pages 16373--16383, 2024.

\bibitem[Kim et~al.(2024)Kim, Pertsch, Karamcheti, Xiao, Balakrishna, Nair, Rafailov, Foster, Lam, Sanketi, et~al.]{kim2024openvla}
Moo~Jin Kim, Karl Pertsch, Siddharth Karamcheti, Ted Xiao, Ashwin Balakrishna, Suraj Nair, Rafael Rafailov, Ethan Foster, Grace Lam, Pannag Sanketi, et~al.
\newblock Openvla: An open-source vision-language-action model.
\newblock \emph{arXiv preprint arXiv:2406.09246}, 2024.

\bibitem[Knoblich et~al.(1999)Knoblich, Ohlsson, Haider, and Rhenius]{8924b34498964e6ca74a2661a102b275}
G{\"u}nther Knoblich, Stellan Ohlsson, Hilde Haider, and Detlef Rhenius.
\newblock Constraint relaxation and chunk decomposition in insight problem solving.
\newblock \emph{Journal of Experimental Psychology: Learning Memory and Cognition}, 25\penalty0 (6):\penalty0 1534--1555, November 1999.
\newblock ISSN 0278-7393.
\newblock \doi{10.1037/0278-7393.25.6.1534}.

\bibitem[Koul(2024)]{Koul2024mazeworld}
Anurag Koul.
\newblock maze-world: Random maze environments with different size and complexity for reinforcement learning and planning research.
\newblock \url{https://koulanurag.dev/maze-world/index.html}, 2024.
\newblock Accessed: 2025-11-07.

\bibitem[Kruger et~al.(2005)Kruger, Carpendale, Scott, and Tang]{10.1145/1054972.1055055}
Russell Kruger, Sheelagh Carpendale, Stacey~D. Scott, and Anthony Tang.
\newblock Fluid integration of rotation and translation.
\newblock In \emph{Proceedings of the SIGCHI Conference on Human Factors in Computing Systems}, CHI '05, page 601–610, New York, NY, USA, 2005. Association for Computing Machinery.
\newblock ISBN 1581139985.
\newblock \doi{10.1145/1054972.1055055}.
\newblock URL \url{https://doi.org/10.1145/1054972.1055055}.

\bibitem[Li et~al.(2024{\natexlab{a}})Li, Ge, Ge, Wang, Wang, Zhang, and Shan]{li2024seed}
Bohao Li, Yuying Ge, Yixiao Ge, Guangzhi Wang, Rui Wang, Ruimao Zhang, and Ying Shan.
\newblock Seed-bench: Benchmarking multimodal large language models.
\newblock In \emph{Proceedings of the IEEE/CVF Conference on Computer Vision and Pattern Recognition}, pages 13299--13308, 2024{\natexlab{a}}.

\bibitem[Li et~al.(2024{\natexlab{b}})Li, Zhang, Wong, Gokmen, Srivastava, Mart{\'\i}n-Mart{\'\i}n, Wang, Levine, Ai, Martinez, et~al.]{li2024behavior}
Chengshu Li, Ruohan Zhang, Josiah Wong, Cem Gokmen, Sanjana Srivastava, Roberto Mart{\'\i}n-Mart{\'\i}n, Chen Wang, Gabrael Levine, Wensi Ai, Benjamin Martinez, et~al.
\newblock Behavior-1k: A human-centered, embodied ai benchmark with 1,000 everyday activities and realistic simulation.
\newblock \emph{arXiv preprint arXiv:2403.09227}, 2024{\natexlab{b}}.

\bibitem[Li et~al.(2022)Li, Li, Xiong, and Hoi]{li2022blip}
Junnan Li, Dongxu Li, Caiming Xiong, and Steven Hoi.
\newblock Blip: Bootstrapping language-image pre-training for unified vision-language understanding and generation.
\newblock In \emph{International conference on machine learning}, pages 12888--12900. PMLR, 2022.

\bibitem[{Lisa Dunlap} et~al.(2025){Lisa Dunlap}, {Trevor Darrell}, {Jacob Steinhardt}, and {Joseph Gonzalez}]{stringsight}
{Lisa Dunlap}, {Trevor Darrell}, {Jacob Steinhardt}, and {Joseph Gonzalez}.
\newblock Stringsight: Automating analysis of model outputs.
\newblock \url{https://www.stringsight.com/}, 2025.

\bibitem[Liu et~al.(2023{\natexlab{a}})Liu, Zhu, Gao, Feng, Liu, Zhu, and Stone]{liu2023libero}
Bo~Liu, Yifeng Zhu, Chongkai Gao, Yihao Feng, Qiang Liu, Yuke Zhu, and Peter Stone.
\newblock Libero: Benchmarking knowledge transfer for lifelong robot learning.
\newblock \emph{Advances in Neural Information Processing Systems}, 36:\penalty0 44776--44791, 2023{\natexlab{a}}.

\bibitem[Liu et~al.(2023{\natexlab{b}})Liu, Li, Wu, and Lee]{liu2023llava}
Haotian Liu, Chunyuan Li, Qingyang Wu, and Yong~Jae Lee.
\newblock Visual instruction tuning.
\newblock In \emph{NeurIPS}, 2023{\natexlab{b}}.

\bibitem[Liu et~al.(2023{\natexlab{c}})Liu, Li, Wu, and Lee]{liu2023visual}
Haotian Liu, Chunyuan Li, Qingyang Wu, and Yong~Jae Lee.
\newblock Visual instruction tuning.
\newblock \emph{Advances in neural information processing systems}, 36:\penalty0 34892--34916, 2023{\natexlab{c}}.

\bibitem[Liu et~al.(2023{\natexlab{d}})Liu, Yu, Zhang, Xu, Lei, Lai, Gu, Ding, Men, Yang, et~al.]{liuagentbench}
Xiao Liu, Hao Yu, Hanchen Zhang, Yifan Xu, Xuanyu Lei, Hanyu Lai, Yu~Gu, Hangliang Ding, Kaiwen Men, Kejuan Yang, et~al.
\newblock Agentbench: Evaluating llms as agents.
\newblock In \emph{The Twelfth International Conference on Learning Representations}, 2023{\natexlab{d}}.

\bibitem[Liu et~al.(2024{\natexlab{a}})Liu, Zhang, Gu, Iong, Xu, Song, Zhang, Lai, Liu, Zhao, et~al.]{liu2024visualagentbench}
Xiao Liu, Tianjie Zhang, Yu~Gu, Iat~Long Iong, Yifan Xu, Xixuan Song, Shudan Zhang, Hanyu Lai, Xinyi Liu, Hanlin Zhao, et~al.
\newblock Visualagentbench: Towards large multimodal models as visual foundation agents.
\newblock \emph{arXiv preprint arXiv:2408.06327}, 2024{\natexlab{a}}.

\bibitem[Liu et~al.(2024{\natexlab{b}})Liu, Duan, Zhang, Li, Zhang, Zhao, Yuan, Wang, He, Liu, et~al.]{liu2024mmbench}
Yuan Liu, Haodong Duan, Yuanhan Zhang, Bo~Li, Songyang Zhang, Wangbo Zhao, Yike Yuan, Jiaqi Wang, Conghui He, Ziwei Liu, et~al.
\newblock Mmbench: Is your multi-modal model an all-around player?
\newblock In \emph{European conference on computer vision}, pages 216--233. Springer, 2024{\natexlab{b}}.

\bibitem[Lu et~al.(2023)Lu, Bansal, Xia, Liu, Li, Hajishirzi, Cheng, Chang, Galley, and Gao]{lu2023mathvista}
Pan Lu, Hritik Bansal, Tony Xia, Jiacheng Liu, Chunyuan Li, Hannaneh Hajishirzi, Hao Cheng, Kai-Wei Chang, Michel Galley, and Jianfeng Gao.
\newblock Mathvista: Evaluating mathematical reasoning of foundation models in visual contexts.
\newblock \emph{arXiv preprint arXiv:2310.02255}, 2023.

\bibitem[Mandlekar et~al.(2021)Mandlekar, Xu, Wong, Nasiriany, Wang, Kulkarni, Fei-Fei, Savarese, Zhu, and Mart{\'\i}n-Mart{\'\i}n]{mandlekar2021matters}
Ajay Mandlekar, Danfei Xu, Josiah Wong, Soroush Nasiriany, Chen Wang, Rohun Kulkarni, Li~Fei-Fei, Silvio Savarese, Yuke Zhu, and Roberto Mart{\'\i}n-Mart{\'\i}n.
\newblock What matters in learning from offline human demonstrations for robot manipulation.
\newblock \emph{arXiv preprint arXiv:2108.03298}, 2021.

\bibitem[McCallum(1994)]{NIPS1994_d2ed45a5}
R.~Andrew McCallum.
\newblock Instance-based state identification for reinforcement learning.
\newblock In G.~Tesauro, D.~Touretzky, and T.~Leen, editors, \emph{Advances in Neural Information Processing Systems}, volume~7. MIT Press, 1994.
\newblock URL \url{https://proceedings.neurips.cc/paper_files/paper/1994/file/d2ed45a52bc0edfa11c2064e9edee8bf-Paper.pdf}.

\bibitem[Mees et~al.(2022)Mees, Hermann, Rosete-Beas, and Burgard]{mees2022calvin}
Oier Mees, Lukas Hermann, Erick Rosete-Beas, and Wolfram Burgard.
\newblock Calvin: A benchmark for language-conditioned policy learning for long-horizon robot manipulation tasks.
\newblock \emph{IEEE Robotics and Automation Letters}, 7\penalty0 (3):\penalty0 7327--7334, 2022.

\bibitem[Michotte(1963)]{michotte1963perception}
Albert Michotte.
\newblock \emph{The Perception of Causality}.
\newblock Routledge, 1 edition, 1963.
\newblock \doi{10.4324/9781315519050}.
\newblock URL \url{https://doi.org/10.4324/9781315519050}.

\bibitem[Misra et~al.(2016)Misra, Zitnick, and Hebert]{misra2016shuffle}
Ishan Misra, C~Lawrence Zitnick, and Martial Hebert.
\newblock Shuffle and learn: unsupervised learning using temporal order verification.
\newblock In \emph{European conference on computer vision}, pages 527--544. Springer, 2016.

\bibitem[Mnih et~al.(2013)Mnih, Kavukcuoglu, Silver, Graves, Antonoglou, Wierstra, and Riedmiller]{mnih2013playing}
Volodymyr Mnih, Koray Kavukcuoglu, David Silver, Alex Graves, Ioannis Antonoglou, Daan Wierstra, and Martin Riedmiller.
\newblock Playing atari with deep reinforcement learning.
\newblock \emph{arXiv preprint arXiv:1312.5602}, 2013.

\bibitem[Niu et~al.(2024)Niu, Sharma, Biamby, Quenum, Bai, Shi, Darrell, and Herzig]{niu2024llarva}
Dantong Niu, Yuvan Sharma, Giscard Biamby, Jerome Quenum, Yutong Bai, Baifeng Shi, Trevor Darrell, and Roei Herzig.
\newblock Llarva: Vision-action instruction tuning enhances robot learning.
\newblock \emph{arXiv preprint arXiv:2406.11815}, 2024.

\bibitem[{Octo Model Team} et~al.(2024){Octo Model Team}, Ghosh, Walke, Pertsch, Black, Mees, Dasari, Hejna, Xu, Luo, Kreiman, Tan, Chen, Sanketi, Vuong, Xiao, Sadigh, Finn, and Levine]{octo_2023}
{Octo Model Team}, Dibya Ghosh, Homer Walke, Karl Pertsch, Kevin Black, Oier Mees, Sudeep Dasari, Joey Hejna, Charles Xu, Jianlan Luo, Tobias Kreiman, {You Liang} Tan, Lawrence~Yunliang Chen, Pannag Sanketi, Quan Vuong, Ted Xiao, Dorsa Sadigh, Chelsea Finn, and Sergey Levine.
\newblock Octo: An open-source generalist robot policy.
\newblock In \emph{Proceedings of Robotics: Science and Systems}, Delft, Netherlands, 2024.

\bibitem[OpenAI(2025)]{OpenAI2025_GPT5SystemCard}
OpenAI.
\newblock {GPT-5 System Card}.
\newblock System card, OpenAI, August 2025.
\newblock URL \url{https://cdn.openai.com/gpt-5-system-card.pdf}.
\newblock Accessed: 2025-11-06.

\bibitem[Qin et~al.(2025)Qin, Ye, Fang, Wang, Liang, Tian, Zhang, Li, Li, Huang, et~al.]{qin2025ui}
Yujia Qin, Yining Ye, Junjie Fang, Haoming Wang, Shihao Liang, Shizuo Tian, Junda Zhang, Jiahao Li, Yunxin Li, Shijue Huang, et~al.
\newblock Ui-tars: Pioneering automated gui interaction with native agents.
\newblock \emph{arXiv preprint arXiv:2501.12326}, 2025.

\bibitem[Ramakrishnan et~al.(2024)Ramakrishnan, Wijmans, Kraehenbuehl, and Koltun]{ramakrishnan2024does}
Santhosh~Kumar Ramakrishnan, Erik Wijmans, Philipp Kraehenbuehl, and Vladlen Koltun.
\newblock Does spatial cognition emerge in frontier models?
\newblock \emph{arXiv preprint arXiv:2410.06468}, 2024.

\bibitem[Ross et~al.(2011)Ross, Gordon, and Bagnell]{ross2011reduction}
St{\'e}phane Ross, Geoffrey Gordon, and Drew Bagnell.
\newblock A reduction of imitation learning and structured prediction to no-regret online learning.
\newblock In \emph{Proceedings of the fourteenth international conference on artificial intelligence and statistics}, pages 627--635. JMLR Workshop and Conference Proceedings, 2011.

\bibitem[Ruoss et~al.(2024)Ruoss, Pardo, Chan, Li, Mnih, and Genewein]{ruoss2024lmact}
Anian Ruoss, Fabio Pardo, Harris Chan, Bonnie Li, Volodymyr Mnih, and Tim Genewein.
\newblock Lmact: A benchmark for in-context imitation learning with long multimodal demonstrations.
\newblock \emph{arXiv preprint arXiv:2412.01441}, 2024.

\bibitem[Sharma et~al.(2024)Sharma, Saxon, and Wang]{sharma-etal-2024-losing}
Aditya Sharma, Michael Saxon, and William~Yang Wang.
\newblock Losing visual needles in image haystacks: Vision language models are easily distracted in short and long contexts.
\newblock In Yaser Al-Onaizan, Mohit Bansal, and Yun-Nung Chen, editors, \emph{Findings of the Association for Computational Linguistics: EMNLP 2024}, pages 5429--5451, Miami, Florida, USA, November 2024. Association for Computational Linguistics.
\newblock \doi{10.18653/v1/2024.findings-emnlp.312}.
\newblock URL \url{https://aclanthology.org/2024.findings-emnlp.312/}.

\bibitem[Shepard and Metzler(1971)]{ShepardMetzler1971}
R.~N. Shepard and J.~Metzler.
\newblock Mental rotation of three‐dimensional objects.
\newblock \emph{Science}, 171\penalty0 (3972):\penalty0 701--703, 1971.
\newblock \doi{10.1126/science.171.3972.701}.

\bibitem[Shi et~al.(2025{\natexlab{a}})Shi, Yang, Liu, Bu, Chen, Zhou, Ma, Wen, Wang, He, et~al.]{shi2025korgym}
Jiajun Shi, Jian Yang, Jiaheng Liu, Xingyuan Bu, Jiangjie Chen, Junting Zhou, Kaijing Ma, Zhoufutu Wen, Bingli Wang, Yancheng He, et~al.
\newblock Korgym: A dynamic game platform for llm reasoning evaluation.
\newblock \emph{arXiv preprint arXiv:2505.14552}, 2025{\natexlab{a}}.

\bibitem[Shi et~al.(2025{\natexlab{b}})Shi, Chen, Chen, Lu, Liu, Ren, Luo, Huang, Yao, and Li]{shi2025diversity}
Modi Shi, Li~Chen, Jin Chen, Yuxiang Lu, Chiming Liu, Guanghui Ren, Ping Luo, Di~Huang, Maoqing Yao, and Hongyang Li.
\newblock Is diversity all you need for scalable robotic manipulation?
\newblock \emph{arXiv preprint arXiv:2507.06219}, 2025{\natexlab{b}}.

\bibitem[Shridhar et~al.(2020)Shridhar, Thomason, Gordon, Bisk, Han, Mottaghi, Zettlemoyer, and Fox]{shridhar2020alfred}
Mohit Shridhar, Jesse Thomason, Daniel Gordon, Yonatan Bisk, Winson Han, Roozbeh Mottaghi, Luke Zettlemoyer, and Dieter Fox.
\newblock Alfred: A benchmark for interpreting grounded instructions for everyday tasks.
\newblock In \emph{Proceedings of the IEEE/CVF conference on computer vision and pattern recognition}, pages 10740--10749, 2020.

\bibitem[Sirdeshmukh et~al.(2025)Sirdeshmukh, Deshpande, Mols, Jin, Cardona, Lee, Kritz, Primack, Yue, and Xing]{sirdeshmukh2025multichallenge}
Ved Sirdeshmukh, Kaustubh Deshpande, Johannes Mols, Lifeng Jin, Ed-Yeremai Cardona, Dean Lee, Jeremy Kritz, Willow Primack, Summer Yue, and Chen Xing.
\newblock Multichallenge: A realistic multi-turn conversation evaluation benchmark challenging to frontier llms.
\newblock \emph{arXiv preprint arXiv:2501.17399}, 2025.

\bibitem[Spaans(2009)]{Spaans2009}
Ruben Spaans.
\newblock Solving sliding-block puzzles.
\newblock Specialisation project report, Norwegian University of Science and Technology (NTNU), 2009.
\newblock URL \url{https://www.pvv.org/~spaans/spec-cs.pdf}.

\bibitem[Srivastava et~al.(2022)Srivastava, Li, Lingelbach, Mart{\'\i}n-Mart{\'\i}n, Xia, Vainio, Lian, Gokmen, Buch, Liu, et~al.]{srivastava2022behavior}
Sanjana Srivastava, Chengshu Li, Michael Lingelbach, Roberto Mart{\'\i}n-Mart{\'\i}n, Fei Xia, Kent~Elliott Vainio, Zheng Lian, Cem Gokmen, Shyamal Buch, Karen Liu, et~al.
\newblock Behavior: Benchmark for everyday household activities in virtual, interactive, and ecological environments.
\newblock In \emph{Conference on robot learning}, pages 477--490. PMLR, 2022.

\bibitem[Stojanovski et~al.(2025)Stojanovski, Stanley, Sharratt, Jones, Adefioye, Kaddour, and K{\"o}pf]{stojanovski2025reasoning}
Zafir Stojanovski, Oliver Stanley, Joe Sharratt, Richard Jones, Abdulhakeem Adefioye, Jean Kaddour, and Andreas K{\"o}pf.
\newblock Reasoning gym: Reasoning environments for reinforcement learning with verifiable rewards.
\newblock \emph{arXiv preprint arXiv:2505.24760}, 2025.

\bibitem[Szot et~al.(2021)Szot, Clegg, Undersander, Wijmans, Zhao, Turner, Maestre, Mukadam, Chaplot, Maksymets, et~al.]{szot2021habitat}
Andrew Szot, Alexander Clegg, Eric Undersander, Erik Wijmans, Yili Zhao, John Turner, Noah Maestre, Mustafa Mukadam, Devendra~Singh Chaplot, Oleksandr Maksymets, et~al.
\newblock Habitat 2.0: Training home assistants to rearrange their habitat.
\newblock \emph{Advances in neural information processing systems}, 34:\penalty0 251--266, 2021.

\bibitem[Team(2025)]{team2025claude}
Claude Team.
\newblock Claude 4 sonnet, 2025.
\newblock URL \url{https://www.anthropic.com/claude/sonnet}.

\bibitem[team(2025)]{gemini3}
Gemini team.
\newblock A new era of intelligence with {Gemini} 3, November 2025.
\newblock URL \url{https://blog.google/products-and-platforms/products/gemini/gemini-3/}.
\newblock Accessed: 2026-01-23.

\bibitem[Team et~al.(2023)Team, Anil, Borgeaud, Alayrac, Yu, Soricut, Schalkwyk, Dai, Hauth, Millican, et~al.]{team2023gemini}
Gemini Team, Rohan Anil, Sebastian Borgeaud, Jean-Baptiste Alayrac, Jiahui Yu, Radu Soricut, Johan Schalkwyk, Andrew~M Dai, Anja Hauth, Katie Millican, et~al.
\newblock Gemini: a family of highly capable multimodal models.
\newblock \emph{arXiv preprint arXiv:2312.11805}, 2023.

\bibitem[Team et~al.(2025{\natexlab{a}})Team, Abeyruwan, Ainslie, Alayrac, Arenas, Armstrong, Balakrishna, Baruch, Bauza, Blokzijl, et~al.]{team2025gemini}
Gemini~Robotics Team, Saminda Abeyruwan, Joshua Ainslie, Jean-Baptiste Alayrac, Montserrat~Gonzalez Arenas, Travis Armstrong, Ashwin Balakrishna, Robert Baruch, Maria Bauza, Michiel Blokzijl, et~al.
\newblock Gemini robotics: Bringing ai into the physical world.
\newblock \emph{arXiv preprint arXiv:2503.20020}, 2025{\natexlab{a}}.

\bibitem[Team et~al.(2025{\natexlab{b}})Team, Kamath, Ferret, Pathak, Vieillard, Merhej, Perrin, Matejovicova, Ram{\'e}, Rivi{\`e}re, et~al.]{team2025gemma}
Gemma Team, Aishwarya Kamath, Johan Ferret, Shreya Pathak, Nino Vieillard, Ramona Merhej, Sarah Perrin, Tatiana Matejovicova, Alexandre Ram{\'e}, Morgane Rivi{\`e}re, et~al.
\newblock Gemma 3 technical report.
\newblock \emph{arXiv preprint arXiv:2503.19786}, 2025{\natexlab{b}}.

\bibitem[Team et~al.(2025{\natexlab{c}})Team, Du, Yin, Xing, Qu, Wang, Chen, Zhang, Du, Wei, et~al.]{team2025kimi}
Kimi Team, Angang Du, Bohong Yin, Bowei Xing, Bowen Qu, Bowen Wang, Cheng Chen, Chenlin Zhang, Chenzhuang Du, Chu Wei, et~al.
\newblock Kimi-vl technical report.
\newblock \emph{arXiv preprint arXiv:2504.07491}, 2025{\natexlab{c}}.

\bibitem[Team et~al.(2025{\natexlab{d}})Team, Barreiros, Beaulieu, Bhat, Cory, Cousineau, Dai, Fang, Hashimoto, Irshad, Itkina, Kuppuswamy, Lee, Liu, McConachie, McMahon, Nishimura, Phillips-Grafflin, Richter, Shah, Srinivasan, Wulfe, Xu, Zhang, Alspach, Angeles, Arora, Guizilini, Castro, Chen, Chu, Creasey, Curtis, Denitto, Dixon, Dusel, Ferreira, Goncalves, Gould, Guoy, Gupta, Han, Hatch, Hathaway, Henry, Hochsztein, Horgan, Iwase, Jackson, Karamcheti, Keh, Masterjohn, Mercat, Miller, Mitiguy, Nguyen, Nimmer, Noguchi, Ong, Onol, Pfannenstiehl, Poyner, Rocha, Richardson, Rodriguez, Seale, Sherman, Smith-Jones, Tago, Tokmakov, Tran, Hoorick, Vasiljevic, Zakharov, Zolotas, Ambrus, Fetzer-Borelli, Burchfiel, Kress-Gazit, Feng, Ford, and Tedrake]{lbmtri2025}
TRI~LBM Team, Jose Barreiros, Andrew Beaulieu, Aditya Bhat, Rick Cory, Eric Cousineau, Hongkai Dai, Ching-Hsin Fang, Kunimatsu Hashimoto, Muhammad~Zubair Irshad, Masha Itkina, Naveen Kuppuswamy, Kuan-Hui Lee, Katherine Liu, Dale McConachie, Ian McMahon, Haruki Nishimura, Calder Phillips-Grafflin, Charles Richter, Paarth Shah, Krishnan Srinivasan, Blake Wulfe, Chen Xu, Mengchao Zhang, Alex Alspach, Maya Angeles, Kushal Arora, Vitor~Campagnolo Guizilini, Alejandro Castro, Dian Chen, Ting-Sheng Chu, Sam Creasey, Sean Curtis, Richard Denitto, Emma Dixon, Eric Dusel, Matthew Ferreira, Aimee Goncalves, Grant Gould, Damrong Guoy, Swati Gupta, Xuchen Han, Kyle Hatch, Brendan Hathaway, Allison Henry, Hillel Hochsztein, Phoebe Horgan, Shun Iwase, Donovon Jackson, Siddharth Karamcheti, Sedrick Keh, Joseph Masterjohn, Jean Mercat, Patrick Miller, Paul Mitiguy, Tony Nguyen, Jeremy Nimmer, Yuki Noguchi, Reko Ong, Aykut Onol, Owen Pfannenstiehl, Richard Poyner, Leticia Priebe~Mendes Rocha, Gordon Richardson, Christopher
  Rodriguez, Derick Seale, Michael Sherman, Mariah Smith-Jones, David Tago, Pavel Tokmakov, Matthew Tran, Basile~Van Hoorick, Igor Vasiljevic, Sergey Zakharov, Mark Zolotas, Rares Ambrus, Kerri Fetzer-Borelli, Benjamin Burchfiel, Hadas Kress-Gazit, Siyuan Feng, Stacie Ford, and Russ Tedrake.
\newblock A careful examination of large behavior models for multitask dexterous manipulation.
\newblock \emph{arXiv preprint arXiv:2507.05331}, 2025{\natexlab{d}}.
\newblock URL \url{https://arxiv.org/abs/2507.05331}.

\bibitem[Todorov et~al.(2012)Todorov, Erez, and Tassa]{todorov2012mujoco}
Emanuel Todorov, Tom Erez, and Yuval Tassa.
\newblock Mujoco: A physics engine for model-based control.
\newblock In \emph{2012 IEEE/RSJ International Conference on Intelligent Robots and Systems}, pages 5026--5033. IEEE, 2012.
\newblock \doi{10.1109/IROS.2012.6386109}.

\bibitem[Touvron et~al.(2023)Touvron, Lavril, Izacard, Martinet, Lachaux, Lacroix, Rozi{\`e}re, Goyal, Hambro, Azhar, et~al.]{touvron2023llama}
Hugo Touvron, Thibaut Lavril, Gautier Izacard, Xavier Martinet, Marie-Anne Lachaux, Timoth{\'e}e Lacroix, Baptiste Rozi{\`e}re, Naman Goyal, Eric Hambro, Faisal Azhar, et~al.
\newblock Llama: Open and efficient foundation language models.
\newblock \emph{arXiv preprint arXiv:2302.13971}, 2023.

\bibitem[Towers et~al.(2024)Towers, Kwiatkowski, Terry, Balis, De~Cola, Deleu, Goul{\~a}o, Kallinteris, Krimmel, KG, et~al.]{towers2024gymnasium}
Mark Towers, Ariel Kwiatkowski, Jordan Terry, John~U Balis, Gianluca De~Cola, Tristan Deleu, Manuel Goul{\~a}o, Andreas Kallinteris, Markus Krimmel, Arjun KG, et~al.
\newblock Gymnasium: A standard interface for reinforcement learning environments.
\newblock \emph{arXiv preprint arXiv:2407.17032}, 2024.

\bibitem[Wang et~al.(2025{\natexlab{a}})Wang, Shi, Tan, Qin, Wang, Zhang, Nambi, Ganu, and Wang]{wang2025multimodal}
Hengyi Wang, Haizhou Shi, Shiwei Tan, Weiyi Qin, Wenyuan Wang, Tunyu Zhang, Akshay Nambi, Tanuja Ganu, and Hao Wang.
\newblock Multimodal needle in a haystack: Benchmarking long-context capability of multimodal large language models.
\newblock In \emph{Proceedings of the 2025 Conference of the Nations of the Americas Chapter of the Association for Computational Linguistics: Human Language Technologies (Volume 1: Long Papers)}, pages 3221--3241, 2025{\natexlab{a}}.

\bibitem[Wang et~al.(2025{\natexlab{b}})Wang, Zhang, Wang, Gao, Li, Wang, Chen, Wan, Lu, Yang, et~al.]{wang2025vagen}
Kangrui Wang, Pingyue Zhang, Zihan Wang, Yaning Gao, Linjie Li, Qineng Wang, Hanyang Chen, Chi Wan, Yiping Lu, Zhengyuan Yang, et~al.
\newblock Vagen: Reinforcing world model reasoning for multi-turn vlm agents.
\newblock \emph{arXiv preprint arXiv:2510.16907}, 2025{\natexlab{b}}.

\bibitem[Wang et~al.(2025{\natexlab{c}})Wang, Zhu, Tang, Li, Xiong, Yu, and Blaschko]{wang2025jigsaw}
Zifu Wang, Junyi Zhu, Bo~Tang, Zhiyu Li, Feiyu Xiong, Jiaqian Yu, and Matthew~B Blaschko.
\newblock Jigsaw-r1: A study of rule-based visual reinforcement learning with jigsaw puzzles.
\newblock \emph{arXiv preprint arXiv:2505.23590}, 2025{\natexlab{c}}.

\bibitem[Wang et~al.(2024)Wang, Xia, He, Chen, Liu, Zhu, Liang, Wu, Liu, Malladi, et~al.]{wang2024charxiv}
Zirui Wang, Mengzhou Xia, Luxi He, Howard Chen, Yitao Liu, Richard Zhu, Kaiqu Liang, Xindi Wu, Haotian Liu, Sadhika Malladi, et~al.
\newblock Charxiv: Charting gaps in realistic chart understanding in multimodal llms.
\newblock \emph{Advances in Neural Information Processing Systems}, 37:\penalty0 113569--113697, 2024.

\bibitem[Wei et~al.(2025)Wei, Sun, Papay, McKinney, Han, Fulford, Chung, Passos, Fedus, and Glaese]{wei2025browsecomp}
Jason Wei, Zhiqing Sun, Spencer Papay, Scott McKinney, Jeffrey Han, Isa Fulford, Hyung~Won Chung, Alex~Tachard Passos, William Fedus, and Amelia Glaese.
\newblock Browsecomp: A simple yet challenging benchmark for browsing agents.
\newblock \emph{arXiv preprint arXiv:2504.12516}, 2025.

\bibitem[Wu et~al.(2024)Wu, Biamby, Quenum, Gupta, Gonzalez, Darrell, and Chan]{wu2024visual}
Tsung-Han Wu, Giscard Biamby, Jerome Quenum, Ritwik Gupta, Joseph~E Gonzalez, Trevor Darrell, and David~M Chan.
\newblock Visual haystacks: A vision-centric needle-in-a-haystack benchmark.
\newblock \emph{arXiv preprint arXiv:2407.13766}, 2024.

\bibitem[xAI(2025)]{xAI2025_Grok4ModelCard}
xAI.
\newblock {Grok 4 Model Card}.
\newblock Model card, xAI, August 2025.
\newblock URL \url{https://data.x.ai/2025-08-20-grok-4-model-card.pdf}.
\newblock Accessed: 2025-11-06.

\bibitem[Xie et~al.(2024)Xie, Zhang, Chen, Li, Zhao, Cao, Hua, Cheng, Shin, Lei, et~al.]{xie2024osworld}
Tianbao Xie, Danyang Zhang, Jixuan Chen, Xiaochuan Li, Siheng Zhao, Ruisheng Cao, Toh~J Hua, Zhoujun Cheng, Dongchan Shin, Fangyu Lei, et~al.
\newblock Osworld: Benchmarking multimodal agents for open-ended tasks in real computer environments.
\newblock \emph{Advances in Neural Information Processing Systems}, 37:\penalty0 52040--52094, 2024.

\bibitem[Yang et~al.(2025{\natexlab{a}})Yang, Li, Yang, Zhang, Hui, Zheng, Yu, Gao, Huang, Lv, et~al.]{yang2025qwen3}
An~Yang, Anfeng Li, Baosong Yang, Beichen Zhang, Binyuan Hui, Bo~Zheng, Bowen Yu, Chang Gao, Chengen Huang, Chenxu Lv, et~al.
\newblock Qwen3 technical report.
\newblock \emph{arXiv preprint arXiv:2505.09388}, 2025{\natexlab{a}}.

\bibitem[Yang et~al.(2025{\natexlab{b}})Yang, Chen, Zhang, Zhao, Qian, Wang, Wang, Koripella, Movahedi, Li, et~al.]{yang2025embodiedbench}
Rui Yang, Hanyang Chen, Junyu Zhang, Mark Zhao, Cheng Qian, Kangrui Wang, Qineng Wang, Teja~Venkat Koripella, Marziyeh Movahedi, Manling Li, et~al.
\newblock Embodiedbench: Comprehensive benchmarking multi-modal large language models for vision-driven embodied agents.
\newblock \emph{arXiv preprint arXiv:2502.09560}, 2025{\natexlab{b}}.

\bibitem[Yu et~al.(2020)Yu, Quillen, He, Julian, Hausman, Finn, and Levine]{yu2020meta}
Tianhe Yu, Deirdre Quillen, Zhanpeng He, Ryan Julian, Karol Hausman, Chelsea Finn, and Sergey Levine.
\newblock Meta-world: A benchmark and evaluation for multi-task and meta reinforcement learning.
\newblock In \emph{Conference on robot learning}, pages 1094--1100. PMLR, 2020.

\bibitem[Yue et~al.(2024)Yue, Ni, Zhang, Zheng, Liu, Zhang, Stevens, Jiang, Ren, Sun, et~al.]{yue2024mmmu}
Xiang Yue, Yuansheng Ni, Kai Zhang, Tianyu Zheng, Ruoqi Liu, Ge~Zhang, Samuel Stevens, Dongfu Jiang, Weiming Ren, Yuxuan Sun, et~al.
\newblock Mmmu: A massive multi-discipline multimodal understanding and reasoning benchmark for expert agi.
\newblock In \emph{Proceedings of the IEEE/CVF Conference on Computer Vision and Pattern Recognition}, pages 9556--9567, 2024.

\bibitem[Yue et~al.(2025)Yue, Zheng, Ni, Wang, Zhang, Tong, Sun, Yu, Zhang, Sun, et~al.]{yue2025mmmu}
Xiang Yue, Tianyu Zheng, Yuansheng Ni, Yubo Wang, Kai Zhang, Shengbang Tong, Yuxuan Sun, Botao Yu, Ge~Zhang, Huan Sun, et~al.
\newblock Mmmu-pro: A more robust multi-discipline multimodal understanding benchmark.
\newblock In \emph{Proceedings of the 63rd Annual Meeting of the Association for Computational Linguistics (Volume 1: Long Papers)}, pages 15134--15186, 2025.

\bibitem[Zhang et~al.(2025{\natexlab{a}})Zhang, Griffiths, Narasimhan, and Press]{zhang2025videogamebench}
Alex~L Zhang, Thomas~L Griffiths, Karthik~R Narasimhan, and Ofir Press.
\newblock Videogamebench: Can vision-language models complete popular video games?
\newblock \emph{arXiv preprint arXiv:2505.18134}, 2025{\natexlab{a}}.

\bibitem[Zhang et~al.(2016)Zhang, Isola, and Efros]{zhang2016colorful}
Richard Zhang, Phillip Isola, and Alexei~A Efros.
\newblock Colorful image colorization.
\newblock In \emph{European conference on computer vision}, pages 649--666. Springer, 2016.

\bibitem[Zhang et~al.(2025{\natexlab{b}})Zhang, Xu, Liu, Yu, Li, Gao, Fei, Yin, Wu, Jiang, et~al.]{zhang2025vlabench}
Shiduo Zhang, Zhe Xu, Peiju Liu, Xiaopeng Yu, Yuan Li, Qinghui Gao, Zhaoye Fei, Zhangyue Yin, Zuxuan Wu, Yu-Gang Jiang, et~al.
\newblock Vlabench: A large-scale benchmark for language-conditioned robotics manipulation with long-horizon reasoning tasks.
\newblock In \emph{Proceedings of the IEEE/CVF International Conference on Computer Vision}, pages 11142--11152, 2025{\natexlab{b}}.

\bibitem[Zheng et~al.(2024)Zheng, Zhang, Zhang, Ye, Luo, Feng, and Ma]{zheng2024llamafactory}
Yaowei Zheng, Richong Zhang, Junhao Zhang, Yanhan Ye, Zheyan Luo, Zhangchi Feng, and Yongqiang Ma.
\newblock Llamafactory: Unified efficient fine-tuning of 100+ language models.
\newblock In \emph{Proceedings of the 62nd Annual Meeting of the Association for Computational Linguistics (Volume 3: System Demonstrations)}, Bangkok, Thailand, 2024. Association for Computational Linguistics.
\newblock URL \url{http://arxiv.org/abs/2403.13372}.

\bibitem[Zhou et~al.(2023)Zhou, Xu, Zhu, Zhou, Lo, Sridhar, Cheng, Ou, Bisk, Fried, et~al.]{zhou2023webarena}
Shuyan Zhou, Frank~F Xu, Hao Zhu, Xuhui Zhou, Robert Lo, Abishek Sridhar, Xianyi Cheng, Tianyue Ou, Yonatan Bisk, Daniel Fried, et~al.
\newblock Webarena: A realistic web environment for building autonomous agents.
\newblock \emph{arXiv preprint arXiv:2307.13854}, 2023.

\bibitem[Zhou et~al.(2025)Zhou, Zhu, Wen, Shen, and Xu]{zhou2025vision}
Zhongyi Zhou, Yichen Zhu, Junjie Wen, Chaomin Shen, and Yi~Xu.
\newblock Vision-language-action model with open-world embodied reasoning from pretrained knowledge.
\newblock \emph{arXiv preprint arXiv:2505.21906}, 2025.

\bibitem[Zhu et~al.(2025)Zhu, Wang, Chen, Liu, Ye, Gu, Tian, Duan, Su, Shao, et~al.]{zhu2025internvl3}
Jinguo Zhu, Weiyun Wang, Zhe Chen, Zhaoyang Liu, Shenglong Ye, Lixin Gu, Hao Tian, Yuchen Duan, Weijie Su, Jie Shao, et~al.
\newblock Internvl3: Exploring advanced training and test-time recipes for open-source multimodal models.
\newblock \emph{arXiv preprint arXiv:2504.10479}, 2025.

\end{thebibliography}
